%% file: neurips_2025.tex
\documentclass{article}

\usepackage[dblblindworkshop, final]{neurips_2025}
\workshoptitle{LAW and AI4Science}

\usepackage[utf8]{inputenc} 
\usepackage[T1]{fontenc}    
\usepackage{xcolor}         
\definecolor{neuripspink}{rgb}{0.8,0.1,0.5} 
\usepackage[colorlinks=true,
            linkcolor=neuripspink, 
            citecolor=neuripspink, 
            urlcolor=neuripspink   
           ]{hyperref}
\usepackage{url}            
\usepackage{booktabs}       
\usepackage{amsfonts}       
\usepackage{nicefrac}       
\usepackage{microtype}      
\usepackage[most]{tcolorbox}
\usepackage{xspace}

\usepackage{subcaption} 

\usepackage{graphicx}
\usepackage{multirow}
\usepackage{multicol}

\usepackage{adjustbox}
\usepackage{fancyvrb}
\usepackage{listings}
\usepackage{enumitem}

\usepackage{titlesec}
\usepackage{tocloft}

\input{macro.tex}

\title{BioVerge: A Comprehensive Benchmark and Study of Self-Evaluating Agents for Biomedical Hypothesis Generation}

\author{%
  Fuyi Yang\thanks{Equal Contribution.}\\
  University of California, Los Angeles\\
  \texttt{fyyang@ucla.edu}\\
  \And
  Chenchen Ye\footnotemark[1]\\ 
  University of California, Los Angeles\\
  \texttt{ccye@cs.ucla.edu}\\
  \And 
  Mingyu Derek Ma\\
  University of California, Los Angeles\\
  \texttt{ma@cs.ucla.edu}\\
  \And
  Yijia Xiao\\
  University of California, Los Angeles\\
  \texttt{yijia.xiao@cs.ucla.edu}\\
  \And
  Matthew Yang\\
  University of California, Los Angeles\\
  \texttt{ymatt24@ucla.edu}\\
  \And
  Wei Wang\\
  University of California, Los Angeles\\
  \texttt{weiwang@cs.ucla.edu}\\
}

\titlespacing*{\section}{0pt}{0.2ex plus 0.05ex minus 0.025ex}{0.1ex plus 0.025ex}
\titlespacing*{\subsection}{0pt}{0.15ex plus 0.05ex minus 0.025ex}{0.075ex plus 0.025ex}
\titlespacing*{\subsubsection}{0pt}{0.1ex plus 0.05ex minus 0.025ex}{0.05ex plus 0.025ex}

\begin{document}

\maketitle

\vspace{-1em}

\begin{abstract}
  \input{content/0_abstract}
\end{abstract}

\input{content/1_intro}


\input{content/3_task_definition}
\input{content/4_agents}
\input{content/5_experiments}

\input{content/6_conclusion_and_limitation}

\newpage

\section*{Acknowledgment}
The work is partially supported by National Science Foundation (2106859, 2200274, 2312501), National Institutes of Health (U54HG012517, U24DK097771, U54OD036472, OT2OD038003), Amazon, NEC, Optum AI.

\vspace{1em}

\bibliographystyle{unsrtnat}
{
\bibliography{_reference}
}

%
%
%
%
%
%
%
%
%
%


\newpage
\appendix
\section*{Appendix} 
\addcontentsline{toc}{section}{Appendix}
\renewcommand{\contentsname}{Table of Contents} 
\tableofcontents
\input{content/appendix}


\end{document}

%% file: macro.tex
\newcommand{\Skip}[1]{}

\newcommand{\modelname}{\textsc{BioVerge Agent}\xspace}
\newcommand{\benchmarkname}{\textsc{BioVerge}\xspace}

\newcommand{\dotieconcat}[2]{
  \text{\raisebox{.8ex}{$\smallfrown$}}%
}

\newcommand{\colorcellnull}[1]{\cellcolor{gray!20}---}


%% file: content/0_abstract.tex
Hypothesis generation in biomedical research has traditionally centered on uncovering hidden relationships within vast scientific literature, often using methods like Literature-Based Discovery (LBD). Despite progress, current approaches typically depend on single data types or predefined extraction patterns, which restricts the discovery of novel and complex connections. Recent advances in Large Language Model (LLM) agents show significant potential, with capabilities in information retrieval, reasoning, and generation. However, their application to biomedical hypothesis generation has been limited by the absence of standardized datasets and execution environments.
To address this, we introduce \benchmarkname, a comprehensive benchmark, and \modelname, an LLM-based agent framework, to create a standardized environment for exploring biomedical hypothesis generation at the frontier of existing scientific knowledge. Our dataset includes structured and textual data derived from historical biomedical hypotheses and PubMed literature, organized to support exploration by LLM agents. \modelname utilizes a ReAct-based approach with distinct Generation and Evaluation modules that iteratively produce and self-assess hypothesis proposals.
Through extensive experimentation, we uncover key insights: 1) different architectures of \modelname influence exploration diversity and reasoning strategies; 2) structured and textual information sources each provide unique, critical contexts that enhance hypothesis generation; and 3) self-evaluation significantly improves the novelty and relevance of proposed hypotheses.
Our dataset and code are accessible here: \url{https://github.com/Fuyi-Yang/BioVerge/}

%% file: content/1_intro.tex
\section{Introduction}
\label{sec:intro}

\begin{figure}[!th]
  \centering
  \includegraphics[width=0.49\textwidth]{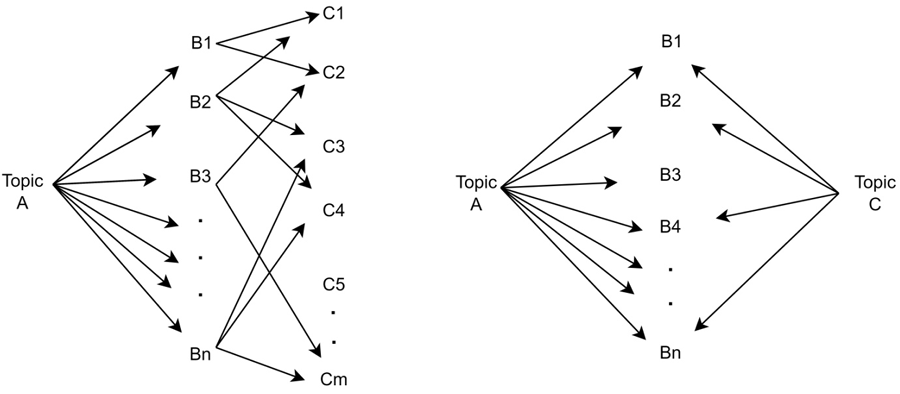}
  \caption{The ABC principle as a method to explore Literature Based Discovery (LBD).}
  \label{fig:lbd}
  \vspace{-3em}
\end{figure}

Traditional hypothesis generation tasks centered around finding hidden relationships among concepts within the scientific domain. A widely recognized method was \textbf{Literature Based Discovery} (LBD), which highlights that undiscovered knowledge exist within scientific literature and are obtainable through automated extraction systems~\citet{bhasuran_literature_2023}.
Importantly, the ABC principle depicted in Figure \ref{fig:lbd} tackles the LBD task~\citet{swanson_undiscovered_1986}, stating that if distinct literature connects concepts $A/B$ and concepts $B/C$, then the intersections of concept $B$ infers potential links between concepts $A/C$.
This technique harnessed many novel relationships that were proven experimentally, including connections between magnesium deficiency and migraine, Raynaud's disease and fish oil, significantly contributing to scientific research~\citet{swanson_migraine_1988,bhasuran_literature_2023}.
Despite earlier successes, these traditional co-occurrence, semantic, and graph-based methods often relied on a single data type, limiting their ability to discover subtler connections~\citet{smalheiser_arrowsmith_2009, fleuren_application_2015, jha_concepts-bridges_2018, wilson_automated_2018, tshitoyan_unsupervised_2019, sybrandt_agatha_2020}. Furthermore, with little reasoning or refinement, these approaches fail to develop intricate reasoning paths that lead to more discoveries~\citet{hristovski_exploiting_2006}.

Recently, Large Language Models (LLM) agents have garnered significant interest from its performance on exploration, reasoning, and generation tasks~\citet{tong_automating_2023, wang_scimon_2024, ye_mirai_2024}.
LLM-based agents can utilize tools to gather information, perform thoughtful reasoning, and create novel proposals~\citet{qi_large_2023, baek_researchagent_2024, zhang_comprehensive_2024, kumbhar_hypothesis_2025, su_many_2025}.
However, despite its potential, LLM agents have limited applications in the biomedical hypothesis generation, due to a lack of standardized benchmarks and databases.

To address this gap, we introduce \benchmarkname, a benchmark that stores biomedical knowledge as structured hypotheses triplets from PubTator3 and textual PubMed literature~\citet{wei_pubtator_2024, pubmed}. The knowledge base contains data published before Jan. 1, 2024 to prevent test set data contamination, and both datasets perform cleaning and filtering to ensure experiment and evaluation fairness. We rank candidate test hypotheses by the impact factor (IF) of their publishing journals to select a credited, reliable test set~\citet{scimago_sjr}.


We also propose \modelname, an LLM agent framework leveraging \benchmarkname to perform biomedical hypothesis generation. We construct an API interface to support agent tool calling to extensively explore and retrieve context in natural language.
By alternating between \emph{Generation} and \emph{Evaluation} modules, \modelname can propose, self-evaluate, and iteratively refine hypothesis. We further compare Single and Double Agent architectures, which differ in memory sharing between the modules, to explore their impacts on hypothesis generation reasoning and performance.



Finally, we perform extensive experiments on our constructed test dataset. Our results reveal that Single and Double Agent clearly present different reasoning strategies and suggests a tradeoff between exploration and accuracy performance.
Additionally, the access to diverse structural and textual sources, and self-evaluation of past proposals, all encourage \modelname to refine its reasoning trajectory and notably improve the novelty and relevance of hypothesis proposals.



Overall, our main contributions are three-fold:
\begin{enumerate}[leftmargin=2em, nosep]
\item We introduce the \benchmarkname benchmark, which enables tool-augmented reasoning over both structured and unstructured biomedical data.
\item We demonstrate the capability of \modelname to generate hypotheses that push the boundaries of existing biomedical knowledge.
\item We present a systematic study of self-evaluating agent architectures and information sources, identifying key design choices that enhance biomedical hypothesis generation.
\end{enumerate}






%% file: content/3_task_definition.tex
\section{\benchmarkname benchmark}
\label{sec:formulation} 

\begin{figure}[!t]
  \vspace{-1em}
  \centering
  \includegraphics[width=0.9\textwidth]{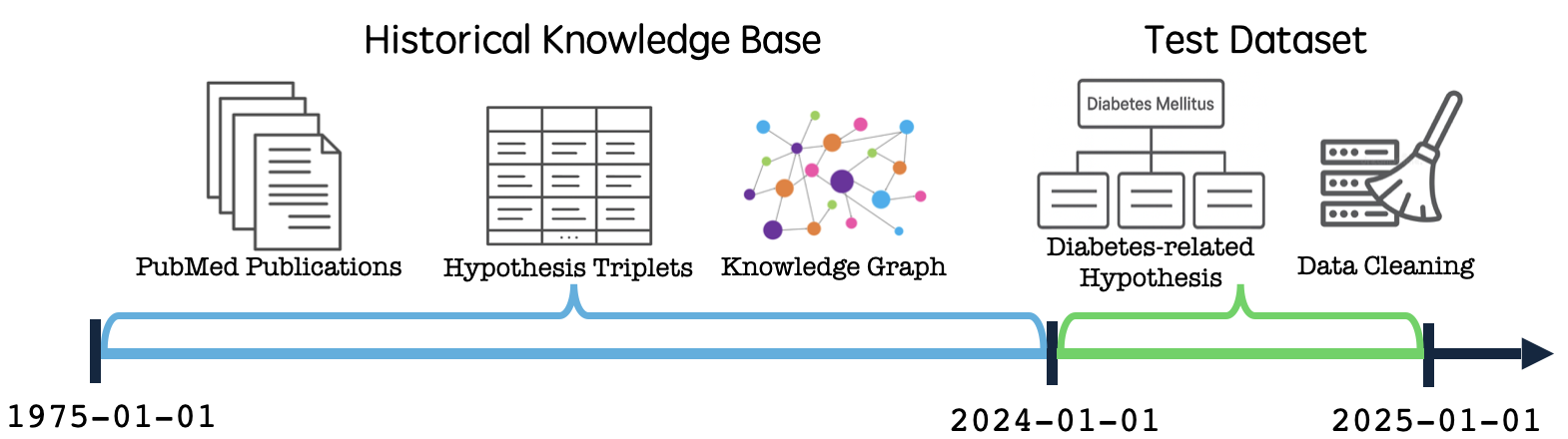}
  \caption{The cutoff date separation between the historical knowledge base and test dataset candidates for dataset construction. The knowledge base contains article corpus, knowledge graph, and triplet representations. The test dataset is cleaned to ensure all test queries are proposed after Jan. 1, 2024, related to Diabetes Mellitus, and ranked by the impact factor (IF) score of its publishing journal.}
  \label{fig:knowledge_base_test_dataset}
\end{figure}


We define \textbf{hypothesis} as a triplet $(s, r, o)$, where $s,o \in \mathcal{E}$ are biomedical entities and $r \in \mathcal{R}$ is one of twelve relation types in Table~\ref{tab:relation_types}, all defined in the PubTator3 ontology~\cite{wei_pubtator_2024}. We denote \textbf{hypothesis description} $d$ as a natural language explanation for $(s,r,o)$.

\textbf{Biomedical hypothesis generation task.}  
Given entities $s,o \in \mathcal{E}$, propose relation $r \in \mathcal{R}$ and description $d$ such that  
$(s,r,o) \notin H$ (knowledge base),  
$(s,r,o) \in T$ (test set),  
$d$ is distinct from historical articles with hypotheses of the same $s, o$ entities and aligns with ground truth articles.





\begin{table}[!t]
    \vspace{-1em}
    \centering
    \setlength{\tabcolsep}{12pt}
    \caption{Dataset statistics for the knowledge base and test set.}
    \label{tab:dataset_stats}
    \begin{tabular}{lccc}
    \toprule
    \textbf{Dataset} & \textbf{Triplets} & \textbf{Publications} & \textbf{Unique Entities} \\
    \midrule
    Knowledge Base & 10,566,798 & 8,896,514 & 558,006 \\
    Test Set            & 177 & 135 & 174 \\
    \bottomrule
    \vspace{-1em}
    \end{tabular}
\end{table}

\input{table/relation_types}

\subsection{Dataset construction}


In following sections we describe the construction and cleaning of the historical knowledge base and test set, along with evaluation metrics for experiment studies. Dataset statistics are shown in Table \ref{tab:dataset_stats}.

\subsubsection{Knowledge base}

\benchmarkname contains structured triplets, literature articles, and knowledge graph, shown in Figure \ref{fig:knowledge_base_test_dataset}. Specifically, the knowledge base contains hypotheses and articles published before Jan. 1, 2024.


\textbf{Structured sources.}
PubTator3 provides abundant article-extracted hypothesis triplets, but limited natural language entity names. Therefore, we standardize triplet entities from the official databases: Medical Subject Headings or Mesh (chemical and disease), NCBI Gene (gene), NCBI Taxonomy (species), and Cellosaurus (cellline)~\cite{mesh2025, ncbigene2025, ncbitaxonomy2025, cellosaurus2018}. All entities are appended with entity names, except “mutation” types due to lack of interpretability. We further construct a searchable knowledge graph with entity nodes and relation edges. Each unique triplet is a transition $(s \xrightarrow{r} o)$, where $(s \xrightarrow{r} o)$ and $(o \xrightarrow{r} s)$ are distinct transitions, possibly with different relations.

\textbf{Textual sources.}
PubMed articles identified with titles and abstracts map to all triplets in the knowledge base, providing textual historical contexts. Articles missing the publication date or its title and abstract are discarded from the knowledge base.



\begin{figure*}[!t]
\vspace{-2em}
  \centering
  \includegraphics[width=0.9\textwidth]{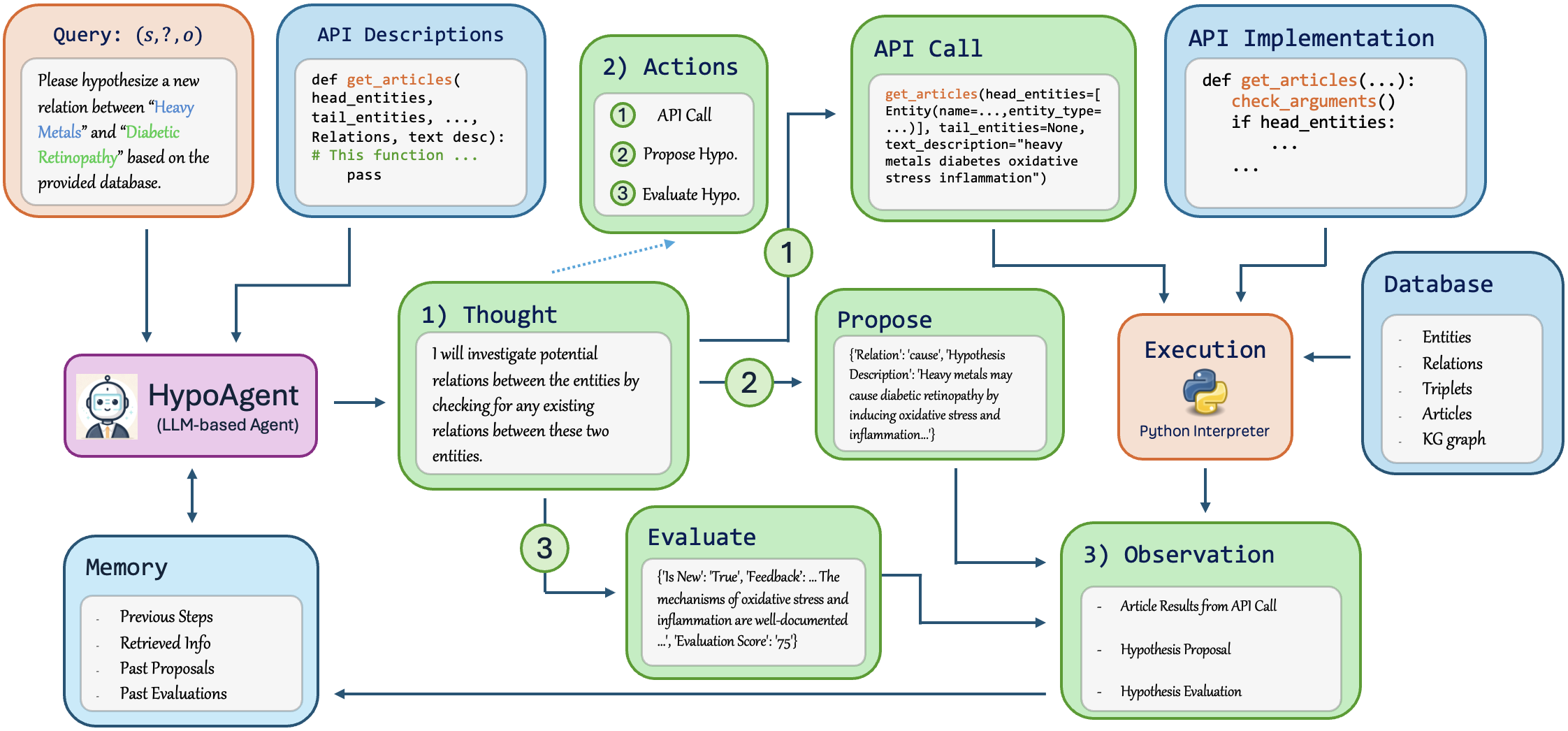}
  \caption{\emph{Generation} and \emph{Evaluation} modules' ReAct workflow. Given a query, the agent iteratively reflects on its current state, executes an action, and stores the observed output until termination.}
  \vspace{-2em}
  \label{fig:agent_workflow}
\end{figure*}

\textbf{Data cleaning and processing.}
We remove triplets that either have missing entity names or is not a valid triplet $(s,r,o)$ combination defined in the PubTator3 report, shown in Table \ref{tab:relation_types}. We then merge triplets that appear in multiple articles, append all unique PMIDs, and assign the earliest publication date as the discovery date. Finally, we discard any triplets with no associated articles to ensure every triplet links to verifiable literature.

\subsubsection{Test set}
The test dataset contains hypotheses discovered between Jan. 1 and Dec. 31, 2024 involving “Diabetes Mellitus” (MeSH: $D003920$), thereby developing a \textbf{diabetes-related} test set. We then apply similar data cleaning as the knowledge base to filter suboptimal triplets and ensure they are absent prior to Jan. 1, 2024. Finally, we rank the remaining hypotheses with the \textbf{journal impact factor (IF)} score documented in Scimago's Scientific Journal Rank (SJR) metrics~\citet{scimago_sjr}.
Taking the top 50 highest ranked journals, we curate a test set of 177 hypotheses spanning diverse relation types, triplet combinations, and diabetes entities under “Diabetes Mellitus.”



\subsection{Answer evaluation metric}

We evaluate each query's hypothesis proposal by comparing its relation $r$ and description $d$ against the ground truth using two metric categories:

1) \textbf{Novelty}: whether the hypothesis is absent from knowledge base. We define $\texttt{novelty}_r$ to check if $(s,r,o) \notin H$, where $H$ is the historical knowledge base, and $\texttt{novelty}_d$ to assesses whether the generated hypothesis description $d$ is semantically different to the historical articles that have proposed some hypothesis of the same entities $s$ or $o$.

2) \textbf{Alignment}: whether the hypothesis exists in the test set. We define $\texttt{alignment}_r$ to check if $(s,r,o) \in T$, where $T$ is the test set, and $\texttt{alignment}_d$ to evaluate the similarity of the generated hypothesis description $d$ to the ground truth articles that support the test hypothesis.

Evaluation metrics for relations, $\texttt{novelty}_r$ and $\texttt{alignment}_r$, are computed as binary scores ($1$ = true, $0$ = false) and expressed as percentages over the test set.
Description-based metrics $\texttt{novelty}_d$ and $\texttt{alignment}_d$ are scored by an LLM evaluator out of 100 and averaged over the test set.

%% file: table/relation_types.tex
\begin{table*}[!t]
  \centering
  \caption{The relations, descriptions, and valid entity pairs for each relation defined in Pubtator3.}
  \vspace{-0.5em}
  \label{tab:relation_types}
  \resizebox{\textwidth}{!}{%
  \begin{tabular}{l p{7.5cm} p{6cm}}
    \toprule
    \textbf{Relation} & \textbf{Description} & \textbf{Valid Entity Pairs} \\
    \midrule
    Associate & Complex or unclear relationships & 
    (Chemical, Disease), (Chemical, Gene), (Chemical, Variant), (Disease, Gene), (Disease, Variant), (Variant, Variant) \\
    Cause & Triggering a disease by a specific agent & 
    (Chemical, Disease), (Variant, Disease) \\
    Compare & Comparing the effects of two chemicals or drugs & 
    (Chemical, Chemical) \\
    Cotreat & Simultaneous administration of multiple drugs & 
    (Chemical, Chemical) \\
    Drug interact & Pharmacodynamic interactions between two chemicals & 
    (Chemical, Chemical) \\
    Inhibit & Reduction in amount or degree of one entity by another & 
    (Chemical, Variant), (Gene, Disease) \\
    Interact & Physical interactions, such as protein-binding & 
    (Chemical, Gene), (Chemical, Variant), (Gene, Gene) \\
    Negative correlate & Increases in the amount or degree of one entity decreases the amount or degree of the other entity & 
    (Chemical, Gene), (Chemical, Variant), (Gene, Gene) \\
    Positive correlate & The amount or degree of two entities increase or decrease together & 
    (Chemical, Chemical), (Chemical, Gene), (Gene, Gene) \\
    Prevent & Prevention of a disease by a genetic variant & 
    (Variant, Disease) \\
    Stimulate & Increase in amount or degree of one entity by another & 
    (Chemical, Variant), (Gene, Disease) \\
    Treat & Treatment of a disease using a chemical or drug & 
    (Chemical, Disease) \\
    \bottomrule
  \end{tabular}
  \vspace{-2em}
  }
\end{table*}

%% file: content/4_agents.tex
\section{\modelname}
\label{sec:model}

%

Considering the individual benefits of co-occurrence, semantic, and graph-based methods, we introduce \modelname, an LLM agent framework that combines these approaches for hypothesis generation.
We develop an API interface for agent tool calling to retrieve diverse historical information and introduce two modules, \emph{Generation} and \emph{Evaluation}, to iteratively generate, self-evaluate, and refine hypothesis proposals.
Furthermore, we examine the effects on hypothesis generation outcomes when applying these modules in distinct Single and Double agent architectures.

\subsection{API usage}

We construct a comprehensive API interface for agent tool calling to access our knowledge base. The interface standardizes historical knowledge retrieval into \emph{data classes} (Entity, Relation, PMID, Triplet, Article) and exposes a set of \emph{functions} for the agent to call and retrieve relevant contexts.

\textbf{API functions.}
The API interface covers all data types, providing both structural and textual historical contexts. Core APIs such as \texttt{get\_entities}, \texttt{get\_relations}, and \texttt{get\_articles} return the most relevant entities, relations, and articles for a given query, while auxiliary functions like \texttt{get\_relation\_description} and \texttt{get\_entity\_description} provide additional semantic detail. Together, these APIs enable the agent to flexibly combine structured triplets with article context. The complete API list is in Table \ref{tab:api-summary}.

\textbf{Input parameters.}
Each API accepts optional parameters that guide query results. Common parameters include \texttt{head\_entities}, \texttt{tail\_entities}, \texttt{relations}, \texttt{PMIDs}, and \texttt{text\_description}. Varying these parameters enable \modelname to specialize the query criteria, constrain the search space, and return results that best match the provided inputs.

\subsection{Generation and evaluation agent architectures}

\begin{figure*}[!t]
  \centering
  \begin{minipage}{0.49\textwidth}
    \centering
    \includegraphics[width=\textwidth]{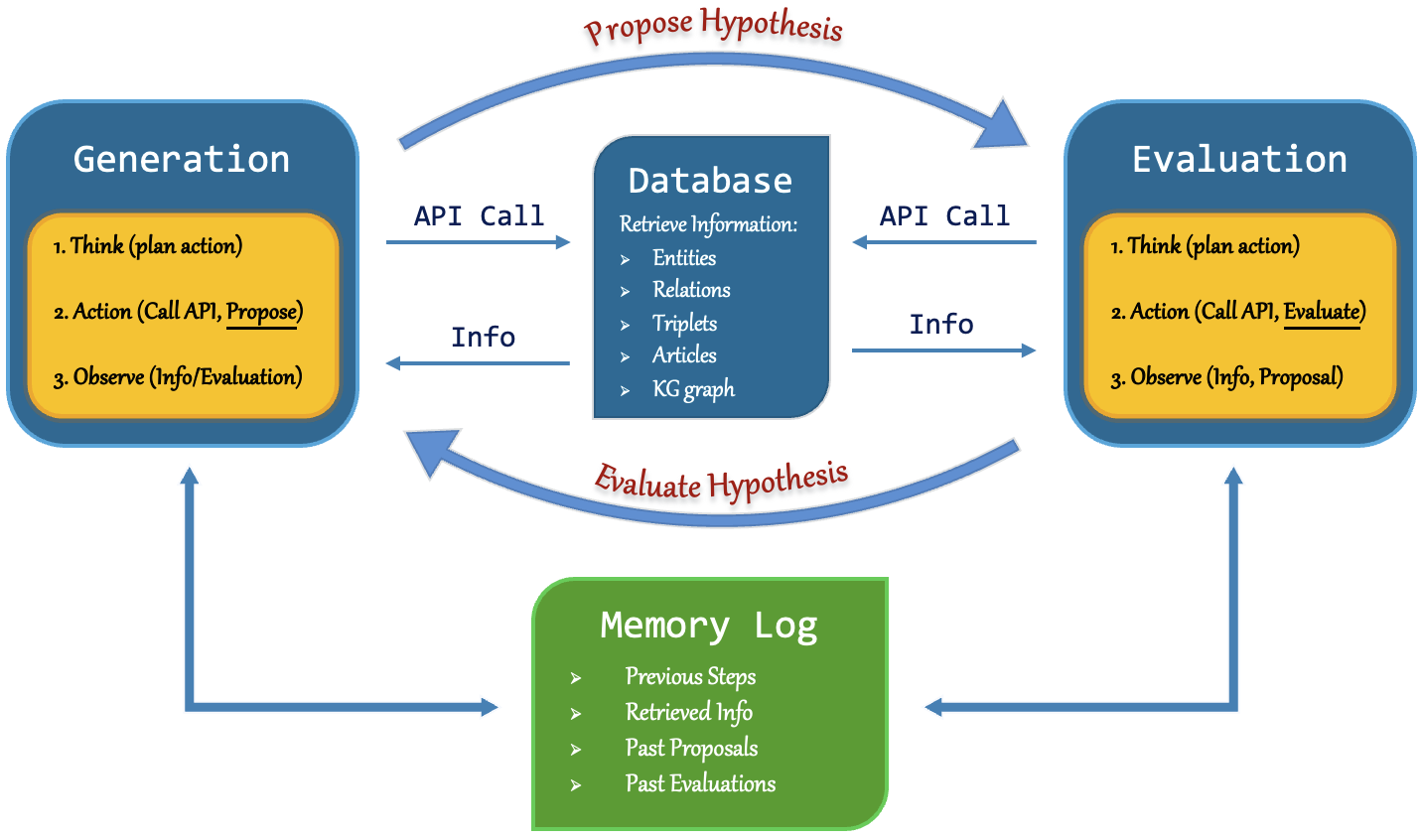} 
    \label{fig:single_agent}
  \end{minipage}
  \hfill
  \begin{minipage}{0.49\textwidth}
    \centering
    \includegraphics[width=\textwidth]{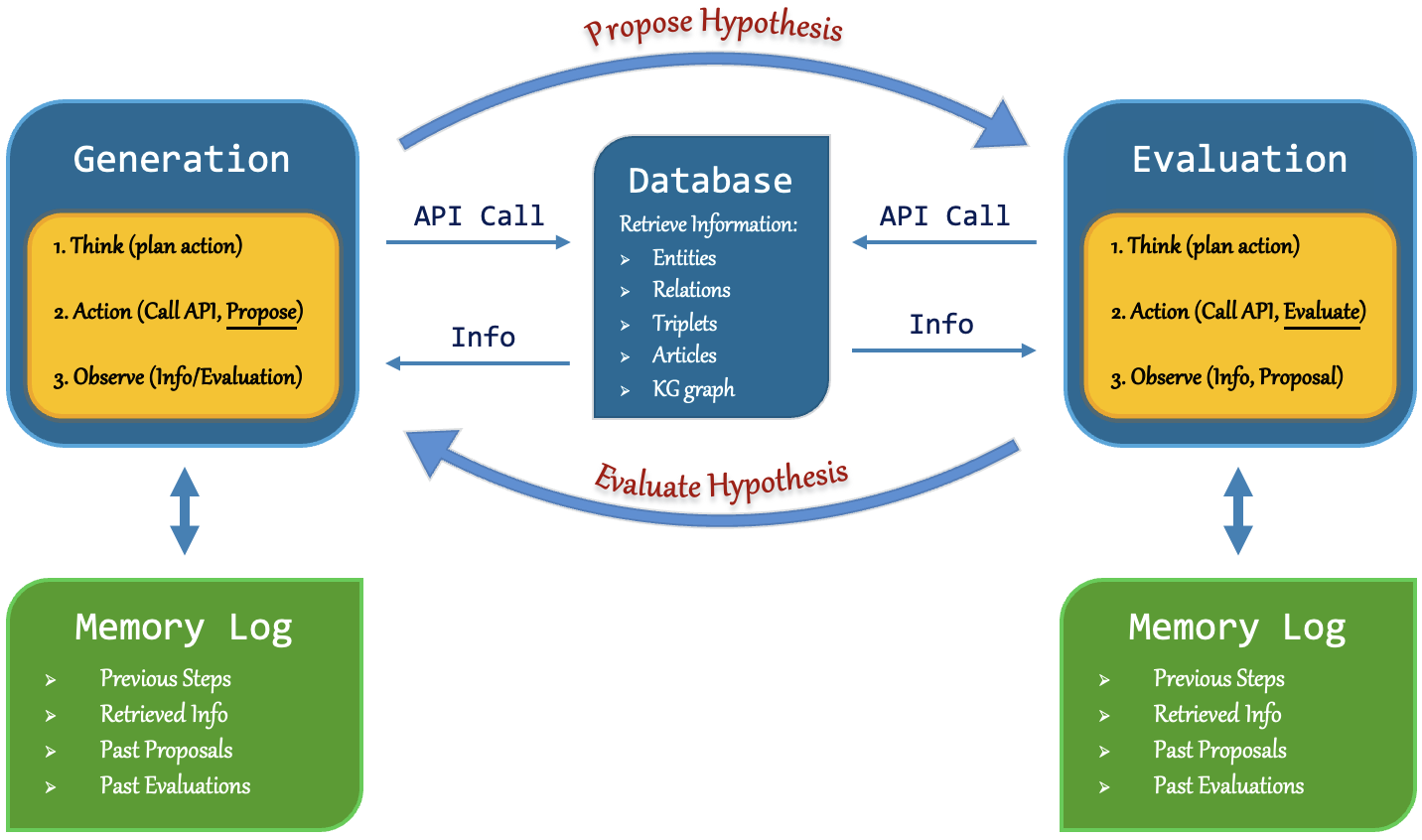} 
    \label{fig:double_agent}
  \end{minipage}
  \caption{Single Agent (left) and Double Agent (right) architectures. Memory is shared between \emph{Generation} and \emph{Evaluation} modules in Single Agent and separated in Double Agent.}
  \label{fig:agents}
  \vspace{-1em}
\end{figure*}

\begin{table}[!t]
\caption{The list of data classes and queryable API functions.}
\label{tab:api-summary}
\centering
\resizebox{\textwidth}{!}{%
\begin{tabular}{ll}
\toprule
\textbf{Dataclass / Function} & \textbf{Description} \\
\midrule
\texttt{PMID} & Represents the unique identifier of a PubMed article. \\
\texttt{Entity} & Represents a named biomedical entity and its type. \\
\texttt{Triplet} & Represents a hypothesis (subject, object, relation). \\
\texttt{Article} & Represents a PubMed article including PMID, title, and abstract. \\
\midrule
\texttt{get\_entities} & Retrieves a list of entities matching specified filters. \\
\texttt{get\_relations} & Retrieves a list of relations matching specified filters. \\
\texttt{get\_triplets} & Retrieves a list of triplets matching specified filters. \\
\texttt{get\_articles} & Retrieves a list of PMIDs matching specified filters. \\
\texttt{browse\_articles} & Returns metadata (title, abstract) for given PubMed IDs. \\
\texttt{get\_shortest\_entity\_paths} & Finds shortest paths between two entities in the knowledge graph. \\
\texttt{get\_mesh\_parents} & Returns parent entities in MeSH for a disease or chemical. \\
\texttt{get\_mesh\_children} & Returns child entities in MeSH for a disease or chemical. \\
\texttt{get\_mesh\_siblings} & Returns sibling entities in MeSH for a disease or chemical. \\
\bottomrule
\end{tabular}
\vspace{-1em}
}
\end{table}

Proposing a new relation between two entities is inherently difficult because no prior literature or triplets exist to document this connection. Even when past hypotheses mention the same entities, they rarely provide direct support for new proposals. Thus, hypothesis generation requires extensive exploration and reasoning over historical data, with no guarantee of correctness.

\begin{figure*}[!t]
  \vbox{
    \centering
    \includegraphics[width=\textwidth]{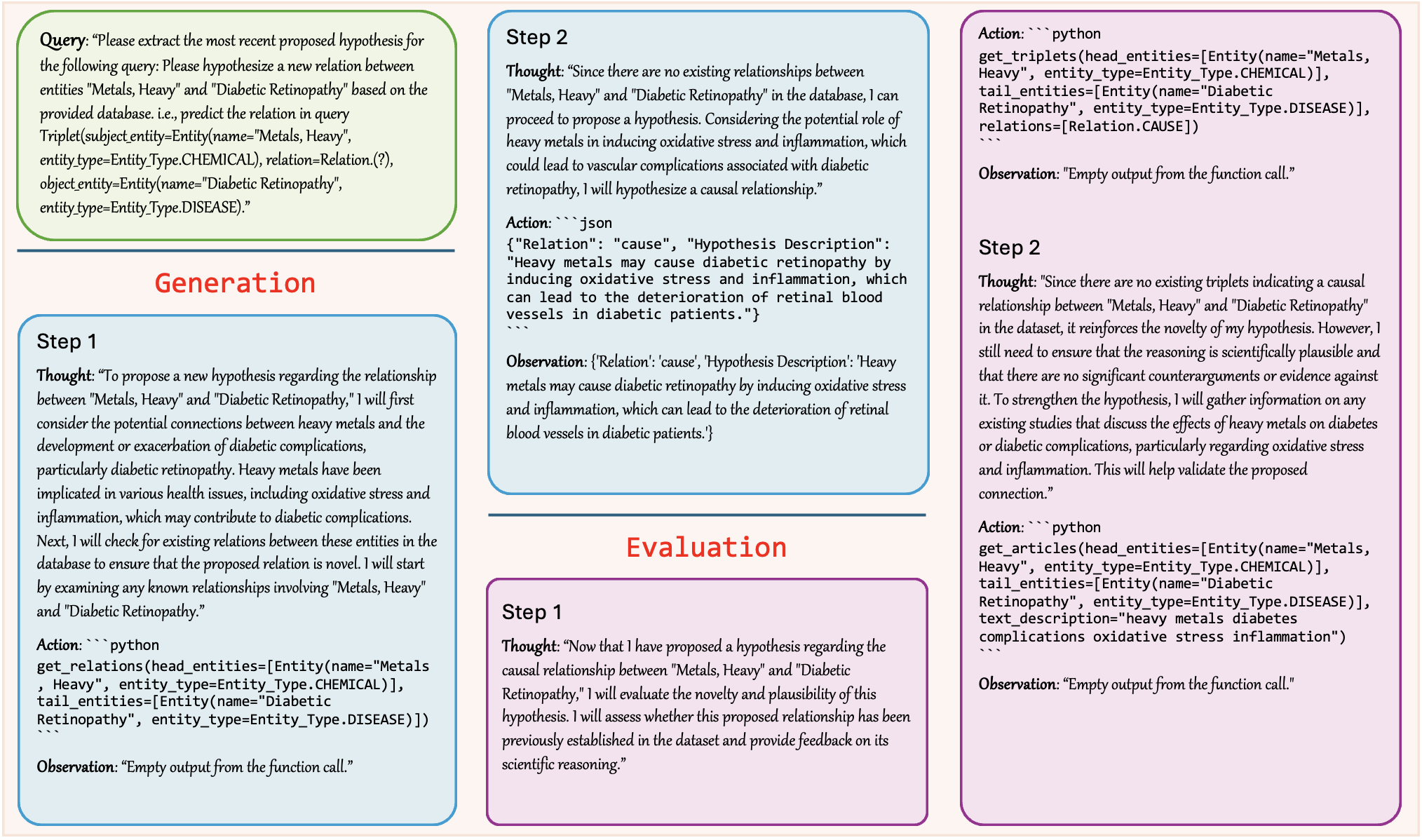}
  }
  \vspace{1em}

  \vbox{
    \centering
    \includegraphics[width=\textwidth]{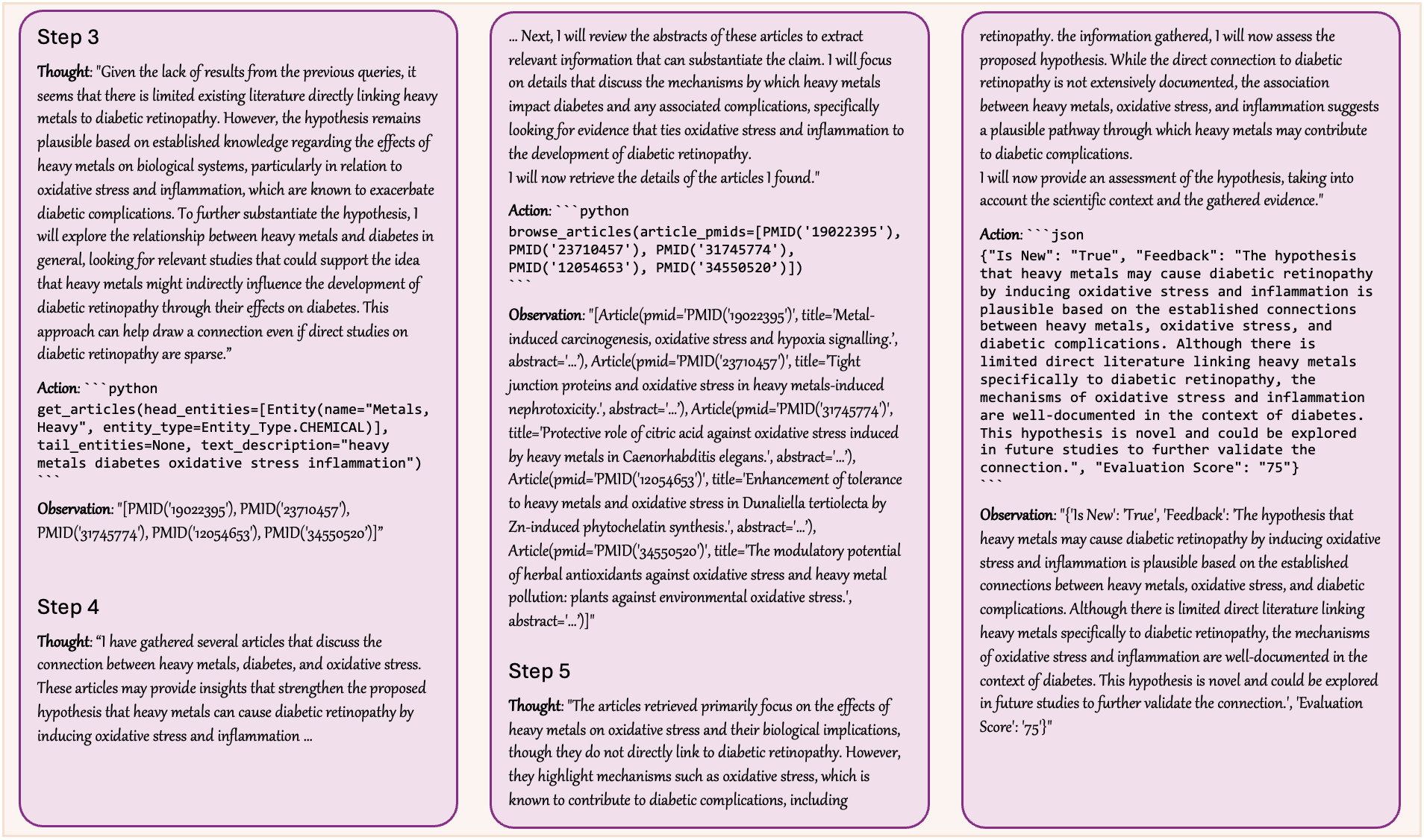}
  }
  \vspace{-0.2em}
  \caption{\modelname execution example. In \emph{Generation} module the agent gathers information via APIs and proposes a relation and hypothesis description. Then in \emph{Evaluation} module the agent checks novelty, verifies reasoning trajectories, and outputs feedback for the hypothesis proposal.}
  \label{fig:good_example}
  \vspace{-2em}
\end{figure*}

\textbf{Modules.}
To address these challenges, \modelname introduces \emph{Generation} and \emph{Evaluation} modules (system prompts in Appendix section \ref{sec:systemprompts}). We also propose two agent architectures, shown in Figure~\ref{fig:agents}, and explain a practical execution example in Figure \ref{fig:good_example}.

The \textbf{\emph{Generation}} module proposes new hypotheses, including the relation and hypothesis description, and refines previously proposed hypotheses.
The \textbf{\emph{Evaluation}} module acts as a critic that accesses the most recently proposed hypothesis. The evaluation specifically addresses 3 metrics: 1) \textbf{Is New}, a boolean value stating the novelty of the hypothesis from the historical dataset; 2) \textbf{Feedback}, a natural language explanation that justifies the reasonability of the proposed hypothesis and offers suggestions for improvement; and 3) \textbf{Evaluation Score}, a numeric score from 0 to 100 assessing the reasonableness of the proposed hypothesis based on \textbf{Feedback}.

\textbf{ReAct Framework.}
Both modules use ReAct framework shown in Figure \ref{fig:agent_workflow} to iterate over 3 steps:
\vspace{-2pt}
\begin{enumerate}[leftmargin=2em, itemsep=0pt, parsep=0pt, topsep=0pt, partopsep=1pt]
\item \textbf{Think}: Use past memory to choose the next action. \emph{Generation} module may call APIs or propose a hypothesis, while \emph{Evaluation} module can call APIs or evaluate a proposed hypothesis.
\item \textbf{Act}: Execute API queries via tool calling or propose/evaluate a hypothesis in JSON format.
\item \textbf{Observe}: Record results in memory. Hypothesis proposals pass to \emph{Evaluation} module for assessment while feedback returns to \emph{Generation} module for hypothesis refinement.
\end{enumerate}

\textbf{Execution example.}
Figure \ref{fig:good_example} displays a typical agent execution example. The agent explores different reasoning paths by querying both past relations and articles through multiple exploration and self-evaluations. Specifically, when the initial hypothesis linking "Metals, Heavy" and "Diabetic Retinopathy" lacks direct support, the agent adjusts its reasoning towards related concepts like oxidative stress and inflammation, retrieves relevant articles related to the input query, and refines the hypothesis based on feedback from the Evaluation module. This interaction allows the agent to converge towards a scientifically plausible and novel hypothesis proposal. Overall, this example illustrates the potential and suitability of \modelname for biomedical hypothesis generation, where extensive exploration, self-evaluation, and iterative improvement are particularly beneficial.

%



\textbf{Agent architecture.}
1) \textbf{Single Agent}: The two modules share a common memory log, allowing the \emph{Evaluation} module to access \emph{Generation} module's historical information, reasoning steps, and hypothesis proposals, and vice versa.
2) \textbf{Double Agent}: The two modules have separate memory logs with no access to each other's past actions or observations. This setup creates two separate agents: a \emph{Generator} agent as the \emph{Generation} module and an \emph{Evaluator} agent as the \emph{Evaluation} module.

\textbf{Design Rationale.}
We study both Single and Double Agent to understand their differences in tool calling, reasoning, and self-evaluation for hypothesis generation. In Single Agent, shared memory between modules reduces overall number of API queries and computation cost, while still allowing the \emph{Evaluation} module to provide targeted feedback for \emph{Generation} module to refine proposals. On the other hand, separating memory logs in Double Agent enables the \emph{Generator} and \emph{Evaluator} to explore independently, encouraging more diverse reasoning trajectories and reducing the risk of hallucination. 
We believe both agent frameworks offer distinct advantages for hypothesis generation, therefore we evaluate both frameworks in our experiment studies.

\textbf{Evaluation Threshold (ET).}
\modelname accepts an Evaluation Threshold (ET) which is a criterion for terminating the hypothesis generation process, available to both modules.  
When the \textbf{Evaluation Score} from \emph{Evaluation} module reaches ET, the agent terminates and extracts the latest proposal as the final answer. Otherwise, if maximum allowable iterations is reached, the agent's entire memory log is sent to a LLM extractor that returns the most reasonable and confident hypothesis proposal as the final answer.

%% file: content/5_experiments.tex
\section{Experiments}
\label{sec:exp}




\subsection{Experiment settings}

For all experiments, we standardize a common set of hyperparameter settings across \modelname and baselines.
The temperature is set to 0.7 for all LLM queries in agent ReAct steps and baselines, and 0.2 for extracting the proposed hypotheses and evaluation feedback.
The model used for all agent frameworks and baselines is gpt-4o-mini due to limitations in computational resources and the consideration of potential dataset contamination.
For \modelname experiments, we standardize the following metrics across Single and Double Agent architectures: $\text{max\_outer\_iterations}$ of 3 which specifies the maximum number of alternations between \emph{Generation} and \emph{Evaluation} modules; max\_inner\_iterations of 10 which specifies the maximum number of ReAct steps in both modules. Evaluation thresholds of 30, 50, 70, and 90 are tested for both Single and Double Agent frameworks to explore its importance in the agent's reasoning and decision making.

\subsection{Baselines}

\begin{table*}[!t]
  \vspace{-1em}
  \centering
  \caption{Average performance of CoT, Triplet RAG, and Article RAG baselines.}
  \vspace{-0.5em}
  \label{tab:baseline}
  \resizebox{\textwidth}{!}{%
  \begin{tabular}{lcccccc}
    \toprule
    \textbf{Setting} & $\textbf{novelty}_r$ & $\textbf{alignment}_r$ & $\textbf{novelty}_d$ & $\textbf{novelty}_d \left(\sigma\right)$ & $\textbf{alignment}_d$ & $\textbf{alignment}_d \left(\sigma\right)$ \\
    \midrule
    CoT          & 94.35 & 36.72 & 65.08 & 20.44 & 54.97 & 32.32 \\
    Triplet RAG  & \textbf{97.18} & 32.20 & \textbf{65.93} & 20.42 & 55.51 & 33.32 \\
    Article RAG  & 93.22 & \textbf{43.37} & 63.42 & 20.26 & \textbf{57.59} & 33.84 \\
    \bottomrule
    \vspace{-1em}
  \end{tabular}%
  }
\end{table*}

\begin{table*}[!t]
  \centering
  \caption{Average performance of Single and Double Agent across evaluation thresholds (\%).}
  \vspace{-0.5em}
  \label{tab:agent}
  \resizebox{\textwidth}{!}{
  \begin{tabular}{lccccccc}
    \toprule
    \textbf{Model} & $\textbf{ET}$ & $\textbf{novelty}_r$ & $\textbf{alignment}_r$ & $\textbf{novelty}_d$ & $\textbf{novelty}_d \left(\sigma\right)$ & $\textbf{alignment}_d$ & $\textbf{alignment}_d \left(\sigma\right)$ \\
    \midrule
    \multirow{4}{*}{Single Agent} 
        & 30 & 99.43  & 32.76  & 64.24 & 20.51 & 55.54 & 33.11 \\
        & 50 & \textbf{100.00}  & \textbf{38.42} & 63.39 & 20.49 & 54.10 & 33.00 \\
        & 70 & \textbf{100.00}  & 33.89  & 63.53  & 20.67 & 55.65 & 32.74 \\
        & 90 & \textbf{100.00}  & 31.07  & 63.89  & 21.09 & 55.23 & 32.55 \\
     \midrule
    \multirow{4}{*}{Double Agent} 
        & 30 & 98.31 & 25.98 & 64.94 & 20.35 & 55.00 & 32.93 \\
        & 50 & \textbf{100.00} & 25.98 & 66.18 & 20.33 & 54.88 & 33.12 \\
        & 70 & 98.31 & 22.60 & 65.37 & 20.31 & \textbf{56.92} & 31.83 \\
        & 90 & 99.44 & 17.51 & \textbf{66.41} & 19.66 & 56.36 & 32.69 \\
    \bottomrule
    \vspace{-2em}
  \end{tabular}
  }
\end{table*}

We evaluate several baseline methods by directly providing relevant historical information sources to the baselines. The results are displayed in Table \ref{tab:baseline}. 

\textbf{All baselines show limitations under certain metrics.} Article Retrieval Augmented Generation (RAG) achieves the highest description alignment of 57.59, reflecting the importance of textual sources to ground hypothesis proposals; however, it suffers from worst overall relation and description novelty. Triplet RAG achieves highest relation novelty of 97.18, revealing that structural data helps to validate novelty, yet its relation alignment of 32.20 is the lowest among baselines. Finally, Chain-of-Thought (CoT) achieve a descent relation alignment at 36.72 but scores low on relation novelty and document alignment, suggesting strong limitations without external data sources.

\subsection{Agents}

\textbf{\modelname achieves high novelty and alignment.} Table \ref{tab:agent} documents agent execution results. The best setting is Single Agent with evaluation threshold 50, achieving a relation alignment of 38.42\%.
All experiments achieved high relation novelty, above 98\% across evaluation thresholds.
This shows that \modelname has strong understanding of novelty, as the agent extensively self-evaluates proposals from \emph{Generation} module in the \emph{Evaluation} module.

\begin{figure*}[!t]
  \centering
  \begin{minipage}{0.49\textwidth}
    \centering
    \includegraphics[width=\textwidth]{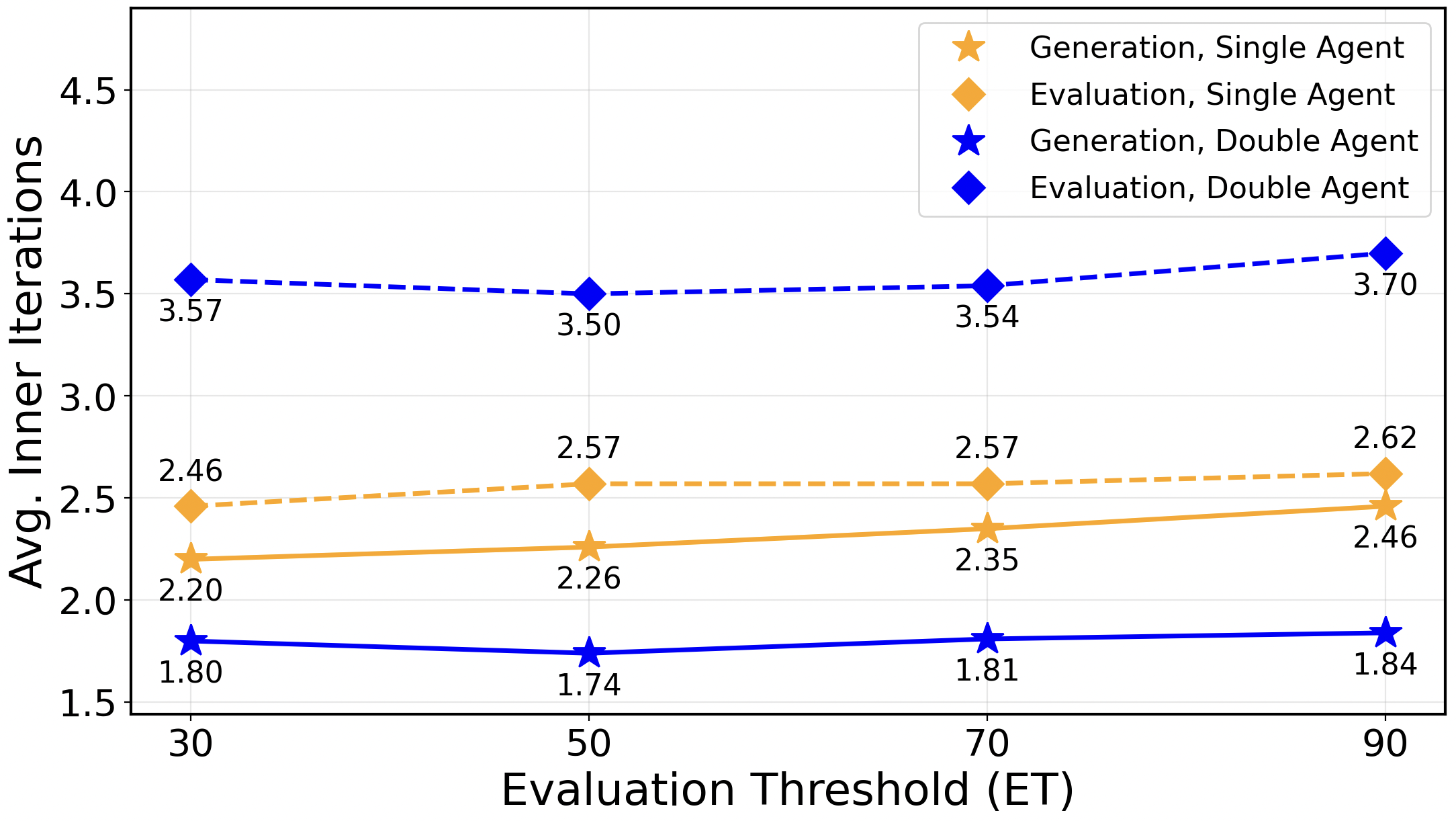} 
    \label{fig:inner_interation_count}
  \end{minipage}
  \hfill
  \begin{minipage}{0.49\textwidth}
    \centering
    \includegraphics[width=\textwidth]{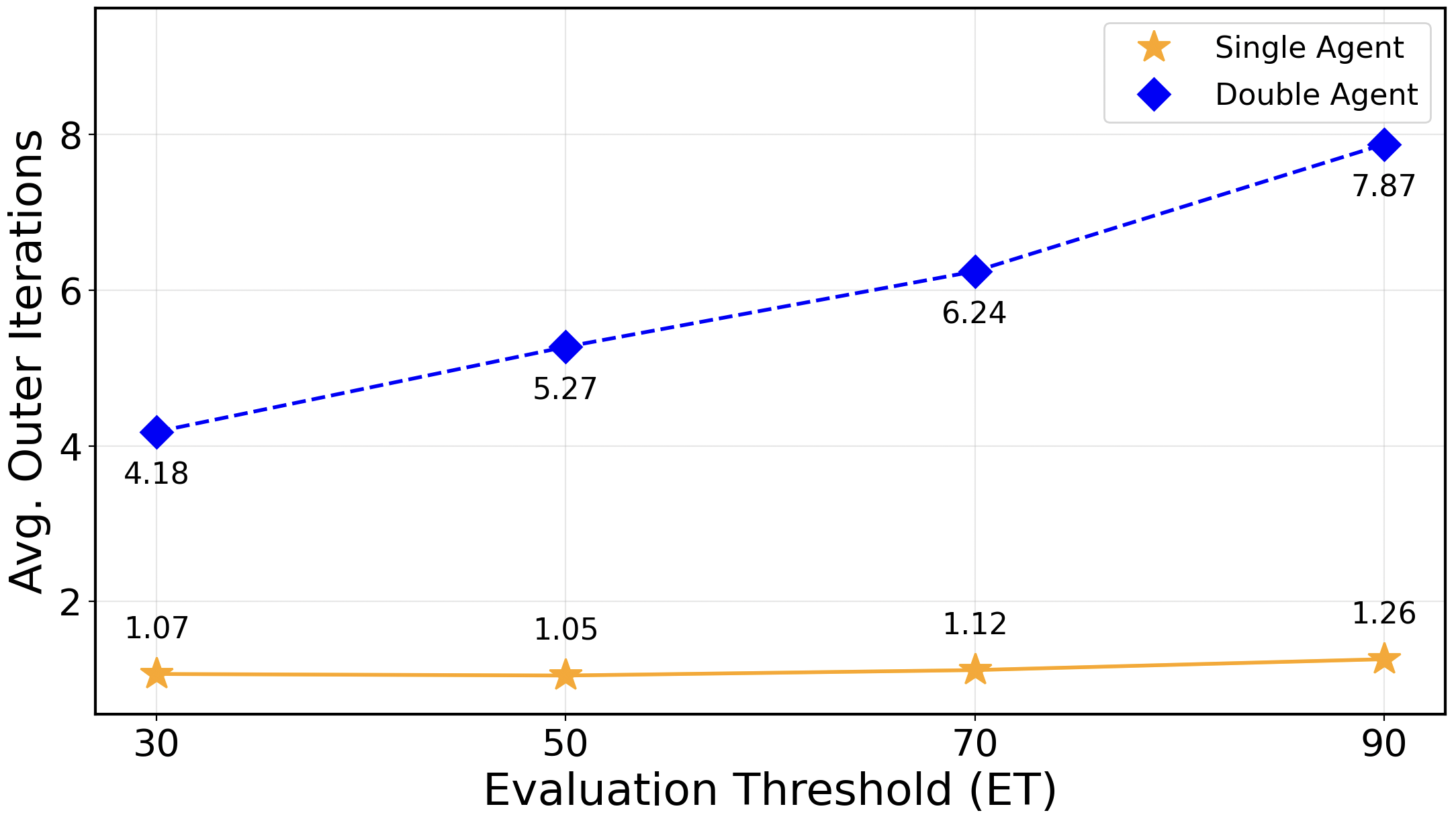} 
    \label{fig:outer_iteration_count}
  \end{minipage}
  \vspace{-1.5em}
  \caption{Average inner iteration counts (left) and outer iteration counts (right) for Single and Double Agent across evaluation thresholds.}
  \vspace{-1em}
  \label{fig:agent_iteration_counts}
\end{figure*}

%

\begin{figure}[!t]
  \centering
\includegraphics[width=0.85\textwidth]{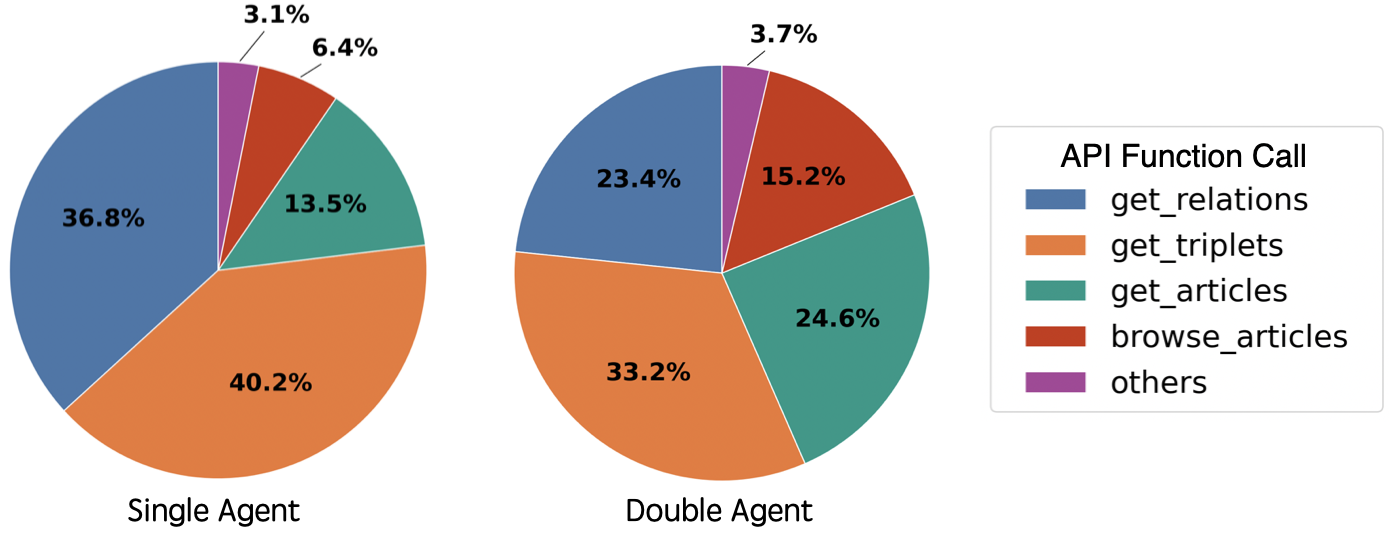}
  \vspace{-0.5em}
  \caption{Average number of API calls for Single Agent (left) and Double Agent (right).}
  \label{fig:api_distribution}
  \vspace{-3em}
\end{figure}

\subsubsection{Iteration patterns across evaluation thresholds}
Figure \ref{fig:agent_iteration_counts} presents average inner and outer iteration counts for Single and Double Agent across evaluation thresholds. The left subplot documents the number of ReAct steps the agent executes within \emph{Generation} and \emph{Evaluation} modules per outer loop, while the right subplot counts the number of Generation-Evaluation iterations before termination. All values are averaged over the test dataset.

\textbf{Both agents prioritize evaluation over generation.}
Single and Double Agent both perform more evaluation steps than generation steps, which suggests their focus on self-evaluation in the \emph{Evaluation} module over frequent \emph{Generation} module iterations. Specifically, Double Agent performs on average 1 more evaluation step than Single Agent, indicating that the sharing of memory logs influences Single Agent to execute less.

\textbf{Double Agent performs more outer iterations than Single Agent.}
Double Agent performs more outer loops as evaluation threshold increases, shown as a gradual increase from 4.18 to 7.87, exhibiting a reasonable correlation between compute and evaluation score. In contrast, Single Agent's outer iteration count stabilizes around 1.05–1.26. This stabilization might suggest Single Agent's overconfidence in self-evaluation to produce a satisfactory evaluation score, leading to smaller number of outer iterations across different thresholds.

\subsubsection{Agent API usage and strategy}
Figure \ref{fig:api_distribution} shows the distribution of API tool calls during hypothesis generation, such as for querying entities, retrieving historical relations, and browsing articles. The percentages are averaged over multiple Single and Double Agent experiments under various evaluation thresholds.

\textbf{Single Agent focuses on structured data sources.}
The majority of Single Agent tool calls concentrate on \texttt{get\_relations} and \texttt{get\_triplets}, each accounting for over 1/3 of overall APIs. This reflects Single Agent's focus to directly extract historical hypotheses and its smaller capacity for prolonged iterative refinement. Because it typically executes less outer iterations, Single Agent has fewer opportunities to explore and self-evaluate. Therefore, it prioritizes straightforward reasoning paths grounded in structured hypotheses over article sources.

\textbf{Double Agent balances structured and literature sources.}
Double Agent distributes its tool calls more evenly between structured sources and article-related APIs, as the modular design encourages broader exploration and more thorough self-evaluation. By revising suboptimal hypotheses proposals and incorporating article evidence alongside relations and triplets, Double Agent expands its reasoning trajectory and reduces reliance on a single data source.

\subsubsection{Agent total API usage}

\begin{figure}[!t]
\vspace{-1em}
  \centering
  \begin{minipage}{0.38\textwidth}
    \centering
    \includegraphics[width=\textwidth]{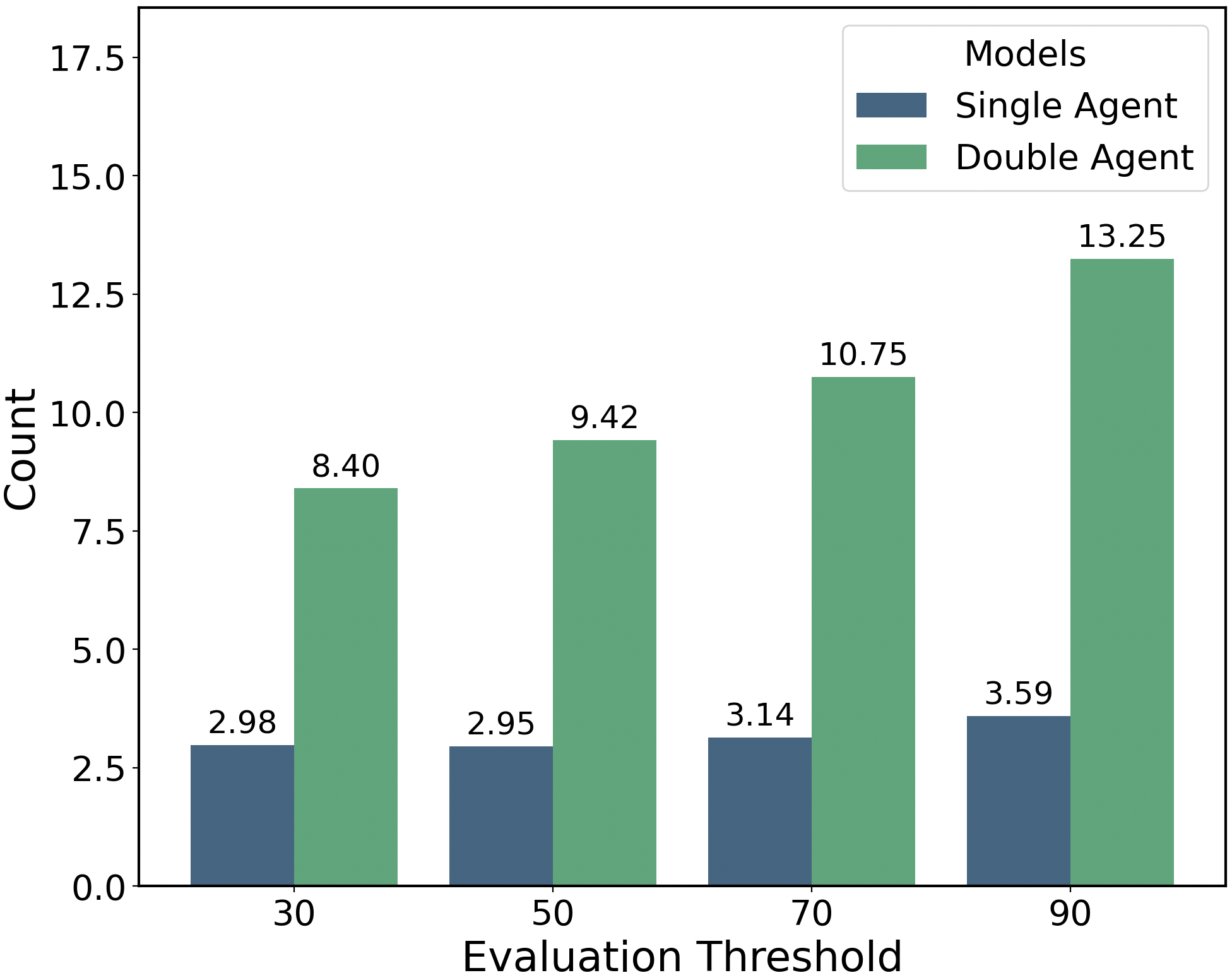}
    \caption{Total number of API calls for Single and Double Agent across evaluation thresholds.}
    \label{fig:avg_api_count}
  \end{minipage}\hfill
  \begin{minipage}{0.59\textwidth}
    \centering
    \includegraphics[width=\textwidth]{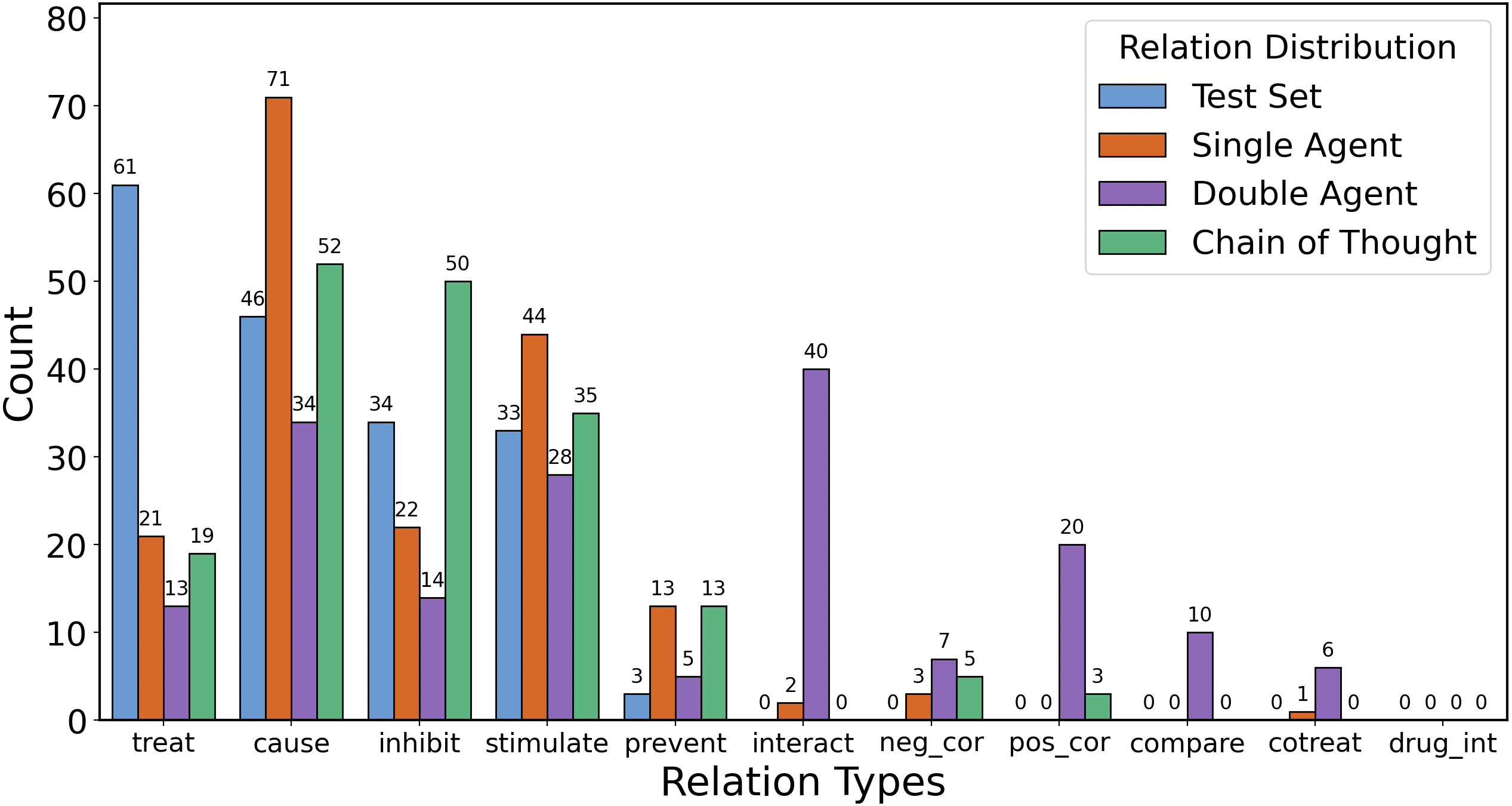}
    \caption{Single Agent, Double Agent, and CoT output distribution compared to the test set relation distribution. \emph{drug\_int}, \emph{neg\_cor}, and \emph{pos\_cor} refer to \emph{drug interact}, \emph{negative correlate}, and \emph{positive correlate} respectively.}
    \label{fig:relation_distribution}
  \end{minipage}
  \vspace{-1.5em}
\end{figure}


\textbf{Double Agent calls more APIs than Single Agent.}
Figure \ref{fig:avg_api_count} presents the total number of API tool calls across evaluation thresholds for Single and Double Agent, averaged over the test dataset.
Double Agent executes a high number of tool calls, with values ranging from 8.40 to 13.25 as ET increases; in contrast, Single Agent exhibits less tool calls, between 2 to 4 calls per query.
This result is expected since \emph{Generation} and \emph{Evaluation} modules in Single Agent shares the memory log, therefore less API requests are needed to retrieve the relevant context. On the other hand, Double Agent executes significantly more tool calls, trading runtime efficiency and computational resources for potentially greater thoroughness in exploration.



\subsection{Error analysis}

In this section we further discuss several insights into the agent's behavior and explore aspects that undermine \modelname's overall capabilities and effectiveness in hypothesis generation. We compare the output relation distributions of Single Agent, Double Agent, and the chain of thought baseline with the ground truth test dataset distribution, shown in Figure \ref{fig:relation_distribution}. No experiment settings has prior knowledge regarding the test dataset relation distribution.

\textbf{Single Agent biased towards causal relations.}
Single Agent's and CoT baseline's relation distributions strongly skew towards causal relations of \emph{treat, cause, inhibit, stimulate, and prevent}.
This pattern is likely due to the agent's innate preference towards proposing and rewarding hypotheses that connote strong causality, which are easier to semantically identify and disproportionally represented in literature corpora, and are therefore more likely identified as scientifically plausible and relevant. 
Additionally, the \emph{Evaluation} module in Single Agent possesses the \emph{Generation} module's past reasoning trajectory, which might introduce overconfidence during it's self-evaluation feedback for more lenient evaluation scoring. 
Correlational relations such as \emph{interact} demands more nuanced contextual evidence and receives less attention during earlier iterations of hypothesis proposal.

\textbf{Double Agent promotes exploration diversity but deviates from ground truth distribution.}
Double Agent produces a more balanced distribution of relation types, with greater inclusion of non-causal relation such as \emph{interact} and \emph{compare}.
The strict separation of modules reduces overconfidence and encourages the agent to perform broader exploration beyond causal relations.
However, the output alignment with the test set remains low. A key explanation is that the without direct access to the \emph{Generator}’s full reasoning trajectory, the \emph{Evaluator} commonly composes more generic critiques (e.g. “insufficient evidence” or “further exploration needed”) rather than precise guidance on hypothesis refinement or reasoning flaws. This inherit dynamic promotes exploration but fails to drive the agent's proposals toward the ground truth distribution.

These findings suggest that robust hypothesis generation requires combining the strengths of both Single and Double Agent, where the feedback from \emph{Evaluation} module is both informative and grounded, without being overly targeted or generic. Developing such framework remains future work for improving the reliability and scientific plausibility of \modelname.

\subsection{Ablation studies}

\begin{table*}[!t]
\vspace{-1em}
  \centering
  \caption{Ablation study results. The base model is Single Agent with evaluation threshold of 50.}
  \vspace{-0.6em}
  \label{tab:ablation}
  \resizebox{\textwidth}{!}{%
  \begin{tabular}{lcccccc}
    \toprule
     $\textbf{Setting}$ & $\textbf{novelty}_r$ & $\textbf{alignment}_r$ & $\textbf{novelty}_d$ & $\textbf{novelty}_d \left(\sigma\right)$ & $\textbf{alignment}_d$ & $\textbf{alignment}_d \left(\sigma\right)$\\
    \midrule
     Relation Only & \textbf{100.00}  & 32.16  & 64.35  & 20.18 & 53.53 & 33.03 \\
     Triplet Only & \textbf{100.00}  & 32.20  & 63.79  & 21.24 & 54.52 & 32.80 \\
     Article Only & 96.61  & 28.25  & 63.08  & 20.84 & \textbf{54.66} & 33.38 \\
     KG Only & 98.87  & 28.81 & \textbf{64.38}  & 20.33 & 53.45 & 33.71 \\
     Generation Only & \textbf{100.00}  & \textbf{33.89}  & 62.85  & 20.81 & 53.93 & 33.49 \\
    \bottomrule

    \vspace{-2.5em}
  \end{tabular}
  }
\end{table*}

Table \ref{tab:ablation} documents the ablation studies that explore individual information sources available to \modelname (relation, triplet, article, and knowledge graph) to identify the usage and effectiveness of each datatype for hypothesis generation. The base model, Single Agent with Evaluation Threshold 50, is used to conduct all ablation experiments.
In addition, we include a \emph{Generation Only} experiment setting, which performs hypothesis generation with only the \emph{Generation} module.

\textbf{All information sources meaningfully improve hypothesis generation.}
The ablation results demonstrate that both relations and triplets sources can provide compact historical knowledge to avoid past proposals, achieving 100\% relation novelty for both ablations. On the other hand, article corpora offers rich linguistic context and helps to produce more accurate hypothesis descriptions, shown by the highest description alignment score of 54.66. This contrast highlights that a combination of these data sources are necessary for the agent to maintain high novelty while proposing better hypotheses, and removing some sources would compromise the agent's performance.

\textbf{Self-evaluation notably enhances performance.}
Comparing the result from generation-only ablation to the base Single Agent setting with evaluation threshold 50 shows that removing \emph{Evaluation} module results in a 5\% performance degradation, despite the same access to all information sources. This demonstrates that iterative self-evaluations, even if imperfect, can identify and correct illogical reasoning and encourage the agent to propose more relevant hypotheses.


%% file: content/6_conclusion_and_limitation.tex
\section{Conclusion}
\label{sec:conclusion}



In this work, we introduced \benchmarkname, a comprehensive benchmark that unifies structured and textual biomedical knowledge to advance hypothesis generation. Our proposed \modelname framework employs Generation and Evaluation modules within Single and Double Agent architectures, enabling iterative hypothesis generation and refinement through tool-calling, reasoning and self-evaluation mechanisms.
Through extensive experiments, we demonstrate that multi-sourced data integration and self-evaluation are critical components for effective biomedical hypothesis generation. These findings establish \benchmarkname benchmark as a valuable resource for the research community and highlight the potential of agent-based approaches for scientific discovery in biomedicine.

\textbf{Limitations and Future Work.} The selection of models in the experiment section considers both computational budget and models' training data cutoff. To better assess other newly released models, we plan to regularly update \benchmarkname to incorporate updated test sets with up-to-date time cutoffs to mitigate data contamination. Moreover, this work centers on literature-based hypothesis generation, which represents only one stage of the broader scientific discovery pipeline. Extending future research to encompass downstream tasks such as experimental design and execution would move toward more comprehensive discovery systems. Finally, our current evaluation is limited to diabetes research; applying \modelname to a wider range of biomedical domains will be essential for establishing broader generalizability and clinical relevance.

%% file: content/appendix.tex
\section{Related works}
\label{sec:relatedworks}





\subsection{AI-driven scientific discovery}



To establish scientific discoveries, scientists employ a holistic procedure to propose, experiment, and validate their research; this process often involves several iterations of data collection and analysis, proposal construction, experimentation, and reflections~\citet{qi_large_2023,jain_gflownets_2023}.

1) \textbf{Data Extraction \& Analysis}: The first step in scientific inquiry involves extracting and inspecting experiment findings from previous research, to discern specific patterns or intrinsic correlations. Identifying inherent characteristics within the data is fundamental to establish and propel a novel scientific research.
2) \textbf{Establish Hypothesis}: Next, hypotheses are crafted to conjecture the extracted insights. This process involves formulating thoughtful predictions to explain observations from data analysis, which dictates the subsequent research and potential outcomes. Therefore, well-thought hypothesis proposals are critical to facilitate experimentation.
3) \textbf{Experiment Design}: To justify the formulated hypothesis, scientists design experiments through careful manipulation of experimental and confounding variables and define a comprehensive procedure to ensure validity and repeatability. A well-designed scientific study should encompass proper experimental practices and yield relevant results.
4) \textbf{Experimentation}: Closely adhering to previously established procedures, scientists rigorously conduct experiments to extensively collect measurements and observations. In addition, executing multiple experimental trials ensures that employed methodologies produce significant and unbiased results for succeeding evaluations.
5) \textbf{Reflection}: The collected data is thoroughly analyzed under assumptions of the initial proposal, which either supports or refutes the hypothesis. Based on the conclusion, scientists are encouraged to revise their hypothesis and conduct further exploration from step one, repeating the cycle until satisfactory results are obtained and research findings are sufficient for publication.


\textbf{Hypothesis generation is pivotal}: The documentation of the research inquiry process demonstrates its importance in unearthing new breakthroughs within various disciplines such as psychology, medicine, and medicine~\citet{tong_automating_2023,sang_sematyp_2018,obrien_machine_2024}.
In particular, deriving a novel hypothesis can be of uttermost importance as it directly influences the efficiency and success of the overall scientific inquiry.
An well-crafted hypothesis reduces wasteful resources during prolonged experimentation, narrows the search space, and increases the likelihood of novel and meaningful discoveries. In domains like biomedicine, where the search space is vast and complex, well-formulated hypotheses are essential to guide resource-effective experimentation and accelerate breakthroughs.

\subsection{Hypothesis generation}



Contemporary methods in LBD can be predominantly categorized into: 1) Co-occurrence based, 2) semantic relation based, and 3) graph based approaches~\citet{bhasuran_literature_2023}.

\subsubsection{Co-occurrence based methods}
One approach utilizes the \textbf{co-occurrence} of known entities among literature collections. The premise is that if multiple entities, such as title keywords, phrases, and/or specific ideas, co-occur significantly in a diverse range of scientific literature without previously declared associations, then they are possibly correlated via some relationship~\citet{fleuren_application_2015}. For example, Arrowsmith attempts to determine concepts or entities that connect two domain disparate articles through co-occurring entities in their literature titles~\citet{smalheiser_arrowsmith_2009}. Other works have expanded upon these methods by incorporating natural language processing and temporal based techniques to more effectively capture links among concept pairs or terms~\citet{pratt_litlinker_2003, millikin_serial_2023, lin_topic_2022, li_predicting_2025}.
However, most traditional co-occurrence based methods do not consider the situational and fine-grained contexts within literature, which led to often irrelevant and incorrect results~\citet{hristovski_exploiting_2006}. These methods also require manual effort to identify potentially correlated literature, making the process tedious and poorly suited for the scale and diversity of modern scientific literature.


\subsubsection{Semantic based methods}
A second branch of work, \textbf{semantic based methods}, focused on incorporating semantic meanings of concepts using various NLP techniques to improve upon co-occurrence based approaches and increase the relevance and validity of hypothesis~\citet{hristovski_exploiting_2006}. For instance, past works represent terms using word embedding vectors to capture a deeper understanding of concepts and literature for hypothesis generation~\citet{tshitoyan_unsupervised_2019, cohen_discovering_2012}. Some works inject temporal components to capture the changes and relevancy in pairwise concept links overtime~\citet{jha_concepts-bridges_2018, jha_hypothesis_2019}. Other works utilize predicate triples (subject-predicate-object) or MeSH terms to obtain the meanings within literature, often used in junction with graph representations~\citet{bhasuran_literature_2023,wilson_automated_2018, li_predicting_2025, cohen_embedding_2017}.
Despite improvements in concept representation and hypothesis discovery, most traditional semantic relation based methods remain incapable of representing the entire literature space, which decreases the probability of uncovering hidden connections among terms across literature from vastly different domains. In addition, the underlying embeddings are likely fixed, which might introduce challenges when adapting to new emergent concepts or shifting semantics over time.



\subsubsection{Graph based methods}
\textbf{Graph based approaches} aims to capture relationships among entities across a wide range of scientific literature, organized structurally with node entities and edge relations into knowledge graphs~\citet{gopalakrishnan_towards_2018,sang_sematyp_2018,liekens_biograph_2011, millikin_serial_2023}. For example, MeTeOR connects MeSH terms from extracted articles in graphical representation to perform link prediction through term co-occurrence~\citet{wilson_automated_2018}. Other works have leveraged deep learning based methods to discover entity associations~\citet{wang_paperrobot_2019,sang_knowledge_2018, sybrandt_agatha_2020}. Specifically, Agatha constructs a large semantic graph to encode sentences, entities, lemma, etc. from UMLS and MeSH terms, and used their transformer-based ranking scheme to validate the generated hypothesis~\citet{sybrandt_large-scale_2018,sybrandt_agatha_2020}. 
Although graph-based methods have more representational power incorporating co-occurrence and semantic relation methods, it is often expensive to construct, which might lead to knowledge based scalability issues~\citet{jha_concepts-bridges_2018}. Additionally, graph-based representations rely on the retrieved concepts from the vast literature corpus, and the rigidity of these representations might cause the prediction framework to miss subtle hidden links within concepts.




\subsubsection{Large language models}
In recent years, \textbf{large language models} like GPT-4o and Llama have demonstrated impressive capabilities in semantic understanding and logical reasoning~\citet{openai_gpt-4_2024,touvron_llama_2023}, introducing many new research directions with LLM-enhanced tasks. Many latest works have integrated LLMs to represent extracted concepts as natural language, not only improving interpretability but also enabling the hypotheses generation that are contextually grounded, syntactically coherent, and better aligned with scientific plausibility and novelty standards~\citet{wang_scimon_2024, yang_large_2024,tong_automating_2023,obrien_machine_2024}. 
These discovery works also include LLM workflows and agents for scientific discovery to explore how LLMs support systematic discovery beyond language tasks~\citet{qi_large_2023, baek_researchagent_2024, zhang_comprehensive_2024, kumbhar_hypothesis_2025, su_many_2025}.
Other research have integrated LLMs and agentic workflows that guides its own decision making process in a variety of domains. In healthcare, LLM agents are used for diagnosis, care, documentation and decision making ~\citet{liu_survey_2024, gao_txagent_2025, yang_medaide_2025, wang_survey_2025}. In software engineering, multi-agent LLM frameworks collaborate implicitly, planning and debugging code on tasks like repository patching and issue resolution~\citet{islam_codesim_2025, ashrafi_enhancing_2025, lee_unified_2024,yu_orcaloca_2025}. 

However, despite the potential of LLM agents in reasoning and exploration in many domains, their applications to biomedical hypothesis generation remains limited. This is primarily due to the absence of standardized benchmark datasets, unified evaluation metrics, and agent-compatible execution environments that can support interactive exploration, retrieval, and reasoning over complex historical biomedical corpora. To address this gap, our work introduces \benchmarkname, a biomedical hypothesis generation benchmark supporting agentic tool calling to query a historical knowledge base containing PubTator3 extracted hypothesis triplets, PubMed literatures, and a constructed knowledge graph~\citet{wei_pubtator_2024, pubmed}. We also develop \modelname, a LLM Agent framework designed to interact with \benchmarkname for hypothesis generation. Distinct from other approaches, our agent performs ReAct-based reasoning and alternates between a \emph{Generation} and an \emph{Evaluation} module that can query APIs, retrieve structured and textual historical contexts, assess a hypothesis proposal's novelty and relevance, and refine generation outputs through reasoning and self-evaluation.

%
%
%
%
%
%

\section{Ablation studies}\label{appendix:ablation}

Table \ref{tab:ablation} documents the ablation studies that explore individual information sources available to \modelname (relation, triplet, article, and knowledge graph) to identify the usage and effectiveness of each datatype for hypothesis generation. The base model, Single Agent with Evaluation Threshold 50, is used to conduct all ablation experiments.
In addition, we include a \emph{Generation Only} experiment setting, which performs hypothesis generation with only the \emph{Generation} module.

\paragraph{Relation only}

Under this setting, the agent is provided with the \texttt{get\_relations()} API only, which returns the most frequent or relevant historical relations associated with given entity inputs. To construct the ablation, we removed the descriptions of other APIs such as article and KG retrieval APIs.
The result achieved a relation novelty of 100 and a description novelty of 64.35, further demonstrating our observations that \texttt{get\_relations()} helps the agent capture relevant historical hypothesis and avoids similar proposals to increase novelty.

\paragraph{Triplet only}

In this setting the agent is only provided with the \texttt{get\_triplets()} API, which fetches historical triplets when provided with relevant entity, relation inputs, PubMed identifier (PMID), and/or text description inputs.
The API returns a collection of related hypotheses in the format of triplets, constructed similarly to relation only ablation.
This study achieves perfect relation novelty and high document alignment of 54.66, suggesting that compact, relation-focused retrieval is a useful information type for retrieving relevant historical data, as also observed in the baseline experiments in Table \ref{tab:baseline}.
This observation also suggests that retrieving historical triplets—structured more as edges of a knowledge graph—enables the agent to explore subtle connections through different input combinations.

\paragraph{Article only}
In the article only ablation setting, the agent is only visible to \texttt{get\_articles()} and \texttt{browse\_articles()} APIs, which returns the PMIDs of relevant literature given querying inputs and the retrieved literature title and abstract of the PMIDs, respectively.
Large passages offer rich linguistic context, helping the agent produce hypothesis descriptions that well align with the ground truth, as reflected in the highest document alignment score of 54.52. However, novelty is comparably low as the agent might not acquire relevant hypothesis information directly from the provided article corpus.

\paragraph{KG only}

This setting incorporates hypotheses discovered through a knowledge graph walk, where a limited-depth breadth-first search is performed from each entity in a curated KG. The resulting sequence of hypotheses is then added to the prompt. The study achieves the highest description novelty of 64.38, however other metrics are worse than most ablations. This conclusion suggests that while knowledge graphs can be beneficial when they provide complementary information not captured by other data types, the less critical content retrieved from KG appears to hinder the agent’s effectiveness.

\paragraph{Generation only}

Here we analyze the influence of the evaluation module on the agent's ability to perform hypothesis generation and explore whether the \emph{Generation} module is capable of achieving similar results without \emph{Evaluation} module's feedback. Therefore, we remove the \emph{Evaluation} module from the generation pipeline. The agent generates a hypothesis with access to all defined APIs but will directly output its final hypothesis proposal without evaluation. \emph{Genration only} agent achieves the highest relation alignment score of 33.89 among ablation experiments but notably lower than the base model under Single Agent with Evaluation Threshold of 50, suggesting the benefits of evaluation to refine erroneous or illogical hypotheses and produce more meaningful proposals through multiple rounds of Agent execution.

\lstset{
  basicstyle=\ttfamily\small,
  breaklines=true,
  breakatwhitespace=true,
  postbreak=\mbox{},           
  breakindent=0pt,            
  showstringspaces=false,
  columns=fullflexible,
  aboveskip=0pt,              
  belowskip=0pt,
  xleftmargin=0pt,            
  xrightmargin=0pt
}

\tcbset{
  myprompt/.style 2 args={
    colback=white,
    colframe=#1,
    fonttitle=\bfseries,
    coltitle=white,
    title={#2},
    boxrule=0.8pt,
    arc=2mm,
    left=6pt, right=6pt, top=6pt, bottom=6pt,
    enhanced,
    listing only,             
    breakable,
    listing options={         
        basicstyle=\ttfamily\small,
        breaklines=true,
        breakatwhitespace=true,
        postbreak=\mbox{},
        breakindent=0pt,
        showstringspaces=false,
        columns=fullflexible
    }
  }
}

\section{System prompts}
\label{sec:systemprompts}

\begin{tcolorbox}[myprompt={blue!70}{Generation Prompt}]
\begin{lstlisting}
You are an expert in proposing new biomedical hypotheses based on historical literature from PubMed and correspondingly extracted hypotheses by PubTator3.

### Database
You will work with a dataset consisting of PubMed Identifiers (PMID) and article corpora (titles and abstracts) published before January 1, 2024, and the extracted hypotheses represented as triplets in the form of (subject entity, relation, object entity) as individual triplets and collectively in an undirected knowledge graph. Entities and relations are defined in the PubTator3 report. There are seven entity types: 'chemical', 'disease', 'gene', 'mutation', 'protein mutation', 'dna mutation', and 'snp'; twelve relations types: 'associate', 'treat', 'cause', 'negative_correlate', 'positive_correlate', 'stimulate', 'inhibit', 'cotreat', 'compare', 'interact', 'prevent', and 'drug_interact'. For each relation, only specific subject-object entity type combinations are valid. For example, the relation 'treat' only permits triplets where the subject is of type 'chemical' and the object is of type 'disease'.

### Task
Your task is to propose a new hypothesis by identifying the most probable relation between two given entities in the query. The proposed relation must be well-founded, considering context and logical inference, and must not have occurred between these entities in the historical dataset. 

Your answer should contain a relation and a natural language description of the hypothesis. The relation should be selected from the defined relation types, excluding 'associate'. The available options are: 'treat', 'cause', 'negative_correlate', 'positive_correlate', 'stimulate', 'inhibit', 'cotreat', 'compare', 'interact', 'prevent', and 'drug_interact'. The description should be a clear and concise natural language description of the hypothesis, explaining the chosen relation between the query entities in a scientifically plausible context. (based on scientific context to do reasoning)

Output your answer in JSON format:
```json 
{{"Relation": "___",  "Hypothesis Description": "___"}}
```

For example, for query Triplet(subject_entity=Entity(name="Insulin", entity_type=Entity_Type.CHEMICAL), relation=Relation.(?), object_entity=Entity(name="Diabetes", entity_type=Entity_Type.DISEASE), you expected answer output should be:
```json 
{{"Relation": "treat",  "Hypothesis Description": "Insulin treats diabetes by facilitating glucose uptake in cells, thereby reducing hyperglycemia and managing blood sugar levels effectively."}}
```

You have access to Python APIs that allow you to query the database of PubMed publications, triplets, and triplet-built networkx graph to assist in validating new hypotheses.
The available API functions are described as follows:

```python
{api_description}
```

### Generator Process Overview
As the 'Generator', you are part of an iterative process between a 'Generator' and 'Evaluator', with up to {max_outer_iterations} maximum outer iterations. Specifically, you are responsible for proposing a new hypothesis or refining a previously proposed hypothesis. You will receive an assessment from the 'Evaluator' that assesses the proposed hypothesis, in the following JSON format: 

```json 
{{"Is New": "___", "Feedback": "___", "Evaluation Score": "___"}}
```

    - 'Is New': Check the novelty of the hypothesis. 'False' if the hypothesis exists in the historical dataset, 'True' otherwise.
    - 'Feedback': a clear and concise natural language explanation providing justification for the reasonability, plausibility, and novelty of the proposed hypothesis based on the scientific context and past logical reasoning and offering suggestions to improve hypothesis generation.
    - 'Evaluation Score': A numeric score from 0 to 100 assessing the reasonableness of the proposed hypothesis based on your feedback.

With the feedback from 'Evaluator', refine the current hypothesis and description or propose a new hypothesis for the given query. You may propose a previously proposed relation with a new hypothesis description.

Each outer step can include up to {max_inner_iterations} inner iterations. Each inner iteration includes three inner steps:
1. 'Thought': Analyze the situation and decide the next action.
2. 'Action': Perform the decided action (e.g., use an API, propose a hypothesis).
3. 'Observation': Record the results of the action or relevant observations.

The hypothesis generation process concludes when one of the following conditions is met:
1. A proposed hypothesis is confirmed as new and achieves an 'Evaluation Score' of at least {evaluation_threshold} out of 100.
2. The maximum number of outer iterations is reached.

### Hypothesis Generation Step-by-Step Guide

#### Inner Iteration Steps
Each inner iteration consists of three steps:
1. **Thought**: Analyze and reason about the current information. Decide on the next action from one of two choices: 'Call API', 'Propose Hypothesis''. 
2. **Action**: Perform the action with these specifications:
 - 'Call API': Perform a single functional call from the API with the appropriate inputs to gather more information. Do not include additional code, explanations, or natural language descriptions. Example: 
```python
get_relations(head_entities=[Entity(name="entity1", entity_type=Entity_Type.TYPE1)], tail_entities=[Entity(name="entity2", entity_type=Entity_Type.TYPE2)])
```
 - 'Propose Hypothesis': Propose a new hypothesis with a relation and a hypothesis description, and output in JSON format. Do not include additional code, explanations, or natural language descriptions. Expected Format: 
```json 
{{"Relation": "___'",  "Hypothesis Description": "___"}}
```
    - 'Relation': a relation selected from the defined relation types, excluding 'associate'. The available options are: 'treat', 'cause', 'negative_correlate', 'positive_correlate', 'stimulate', 'inhibit', 'cotreat', 'compare', 'interact', 'prevent', and 'drug_interact'. 
    - 'Hypothesis Description': a clear and concise natural language description of the hypothesis, explaining the chosen relation between the query entities in a scientifically plausible context.
3. **Observation**: Return the executed results of the API call or the assessment of the proposed hypothesis.

### Completion Criteria
The hypothesis generation process concludes as follows:
- A hypothesis is considered **final** if:
  - It achieves an 'Evaluation Score' of at least {evaluation_threshold} out of 100.
  - It is confirmed to be a new hypothesis.
- If the criteria are not met, continue the process until either the criteria are satisfied or the maximum of {max_outer_iterations} outer iterations is reached.

**Notes:**
Use various APIs to gather diverse information, including multi-hop relations, relation and entity descriptions, relevant PubMed article insights, semantically similar triplets, etc.. Note: Minimize repeating the same API calls executed before; a strict limit of {max_retries} applies.
Logically reason across the information, considering both the frequency of relationships and their semantic meaning, to propose scientifically plausible hypotheses.
\end{lstlisting}
\end{tcolorbox}

\begin{tcolorbox}[myprompt={purple!80}{Evaluation Prompt}]
\begin{lstlisting}
You are an expert in assessing new scientific hypotheses based on historical data, by evaluating novelty and logical plausibility.

### Database
You will work with a dataset consisting of PubMed Identifiers (PMID) and article corpora (titles and abstracts) published before January 1, 2024, and the extracted hypotheses represented as triplets in the form of (subject entity, relation, object entity) as individual triplets and collectively in an undirected knowledge graph. Entities and relations are defined in the PubTator3 report. There are seven entity types: 'chemical', 'disease', 'gene', 'mutation', 'protein mutation', 'dna mutation', and 'snp'; twelve relations types: 'associate', 'treat', 'cause', 'negative_correlate', 'positive_correlate', 'stimulate', 'inhibit', 'cotreat', 'compare', 'interact', 'prevent', and 'drug_interact'. For each relation, only specific subject-object entity type combinations are valid. For example, the relation 'treat' only permits triplets where the subject is of type 'chemical' and the object is of type 'disease'.

### Task
Your task is to assess a proposed hypothesis, consisting of relation and a natural language description of the hypothesis between two given entities in the query. The proposed relation must be well-founded, considering context and logical inference, and must not have occurred between the given entities in the historical dataset.

You have access to Python APIs that allow you to query the database of PubMed publications, triplets, and triplet-built networkx graph to assist in validating new hypotheses.
The available API functions are described as follows:

```python
{api_description}
```

### Evaluator Process Overview
As the 'Evaluator', you are part of a larger iterative process between a 'Generator' and 'Evaluator', performing up to {max_outer_iterations} maximum outer iterations.
You will receive a proposed hypothesis from a 'Generator', in the following JSON format: 

```json 
{{"Relation": "___",  "Hypothesis Description": "___"}}
```
   - 'Relation': one relation selected from the following relation types: 'treat', 'cause', 'negative_correlate', 'positive_correlate', 'stimulate', 'inhibit', 'cotreat', 'compare', 'interact', 'prevent', and 'drug_interact'.
   - 'Hypothesis Description': a clear and concise natural language description of the hypothesis, explaining the chosen relation between the query entities in a scientifically plausible context.

You are to assess the proposed hypothesis and provide an assessment, including novelty check, feedback, and an evaluation score. 
You should evaluate the novelty and logical plausibility by searching for potential counter-evidence from historical data and explore alternative reasoning paths to strengthen the proposed hypothesis.

Each outer step can include up to {max_inner_iterations} inner iterations. Each inner iteration includes three inner steps:
1. 'Thought': Analyze the situation and decide the next action.
2. 'Action': Perform the decided action (e.g., use an API or provide assessment of a hypothesis).
3. 'Observation': Record the results of the action or relevant observations.

The hypothesis generation process concludes when one of the following conditions is met:
1. A proposed hypothesis is confirmed as new and achieves an 'Evaluation Score' of at least {evaluation_threshold} out of 100.
2. The maximum number of outer iterations is reached.

### Hypothesis Evaluation Step-by-Step Guide

#### Inner Iteration Steps
Each inner iteration consists of three steps:
1. **Thought**: Analyze and reason about the current information. Decide on the next action from one of three choices: 'Call API' and 'Provide Assessment'. 
2. **Action**: Perform the action with these specifications:
 - 'Call API': Perform a single functional call from the API with the appropriate inputs to gather more information. Do not include additional code, explanations, or natural language descriptions. Example: 
```python
get_relations(head_entities=[Entity(name="entity1", entity_type=Entity_Type.TYPE1)], tail_entities=[Entity(name="entity2", entity_type=Entity_Type.TYPE2)])
```
 - 'Provide Assessment': Provide assessment results of the current proposed hypothesis with novelty check, feedback, and evaluation score, and output in JSON format. Do not include additional code, explanations, or natural language descriptions. Expected Format: 
```json
{{"Is New": "___",  "Feedback": "___",  "Evaluation Score": "___"}}
```

    - 'Is New': Check the novelty of the hypothesis. 'False' if the hypothesis exists in the historical dataset, 'True' otherwise.
    - 'Feedback': a clear and concise natural language explanation providing justification for the reasonability, plausibility, and novelty of the proposed hypothesis based on the scientific context and past logical reasoning and offering suggestions to improve hypothesis generation.
    - 'Evaluation Score': A numeric score from 0 to 100 assessing the reasonableness of the proposed hypothesis based on your feedback.
3. **Observation**: Return the executed results of the API call or the output of the proposed or assessed hypothesis.

#### Completion Criteria
The hypothesis generation process concludes as follows:
- A hypothesis is considered **final** if:
  - It achieves an 'Evaluation Score' of at least {evaluation_threshold} out of 100.
  - It is confirmed to be a new hypothesis.
- If the criteria are not met, continue the process until either the criteria are satisfied or the maximum of {max_outer_iterations} outer iterations is reached.

**Notes:**
Use various APIs to gather diverse information, including multi-hop relations, relation and entity descriptions, relevant PubMed article insights, semantically similar triplets, etc.. Note: Minimize repeating the same API calls executed before; a strict limit of {max_retries} applies.
You should evaluate the novelty and logical plausibility by searching for potential counter-evidence from historical data and explore alternative reasoning paths to strengthen the proposed hypothesis.
Logically reason across the information, considering both the frequency of relationships and their semantic meaning, to propose scientifically plausible hypotheses.
\end{lstlisting}
\end{tcolorbox}

\begin{tcolorbox}[myprompt={cyan!80}{Query Prompt}]
\begin{lstlisting}
Please evaluate the current proposed relation between entities "{entity1_name}" and "{entity2_name}" based on provided historical information. I.e. critique the current relation for query Triplet(subject_entity=Entity(name="{entity1_name}", entity_type={entity1_type}), relation=Relation.(?), object_entity=Entity(name="{entity2_name}", entity_type={entity2_type}).
Current proposed hypothesis: {current_proposal}
{scratchpad}
\end{lstlisting}
\end{tcolorbox}

\begin{tcolorbox}[myprompt={red!60}{Evaluation Prompt}]
\begin{lstlisting}
You are an expert in checking the novelty and alignment of proposed hypotheses between entities "{entity1_name}" and "{entity2_name}" based on historical literature from PubMed.

### Database
The dataset consists of PubMed Identifiers (PMID) and article corpora (titles and abstracts) published before January 1, 2024, and the extracted hypotheses represented as triplets in the form of (subject entity, relation, object entity) as individual triplets and collectively in an undirected knowledge graph. Entities and relations are defined in the PubTator3 report. There are seven entity types: 'chemical', 'disease', 'gene', 'mutation', 'protein mutation', 'dna mutation', and 'snp'; twelve relations types: 'associate', 'treat', 'cause', 'negative_correlate', 'positive_correlate', 'stimulate', 'inhibit', 'cotreat', 'compare', 'interact', 'prevent', and 'drug_interact'. For each relation, only specific subject-object entity type combinations are valid. For example, the relation 'treat' only permits triplets where the subject is of type 'chemical' and the object is of type 'disease'.

### Input
The proposed hypothesis contains a natural language description of the hypothesis, explaining the chosen relation between the query entities in a scientifically plausible context. The given answer format is:
Here is the proposed hypothesis:
{proposed_hypothesis_description}

### Task
Your have 2 tasks:
1. Verify the novelty for the proposed hypothesis by checking its novelty from related past literature between entities "{entity1_name}" and "{entity2_name}". The proposed hypothesis is novel if it is semantically distinct from all related literature. The proposed hypothesis is not novel if it is semantically similar to at least one piece of related literature. Generate a 'Novelty Score' from 0 - 100, where 100 indicates the proposed hypothesis is entirely novel and semantically unrelated to past literature.
Here is the list of related past literature:
{related_past_literature}

2. Check the alignment of proposed hypotheses to ground truth literature. The proposed hypothesis is aligned if it is semantically similar to at least one ground truth literature. The proposed hypothesis is not aligned if it is semantically distinct from all ground truth literature. Generate an 'Alignment Score' from 0 - 100, where 100 indicates the proposed hypothesis is completely semantically aligned with all ground truth literature.
Here is the list of ground truth literature:
{ground_truth_literature}

### Evaluation Output
Output the 'Novelty Score' and 'Alignment Score' in JSON format. Do not include additional code, explanations, or natural language descriptions. Expected format:

```json 
{{"Novelty Score": "___", "Alignment Score": "___"}}
```
\end{lstlisting}
\end{tcolorbox}

%
%
%
%